\pdfoutput=1

\documentclass{article}

\usepackage[dblblindworkshop, final]{neurips_2025}

\usepackage[utf8]{inputenc} 
\usepackage[T1]{fontenc}    
\usepackage{hyperref}       
\usepackage{url}            
\usepackage{booktabs}       
\usepackage{amsfonts}       
\usepackage{nicefrac}       
\usepackage{microtype}      
\usepackage{xcolor}         

\usepackage{algorithm}
\usepackage{algorithmic}
\usepackage{listings}
\usepackage{xcolor} 
\usepackage{amsmath}

\usepackage{graphicx}

\graphicspath{{plots/}}

\title{Robust LLM Unlearning with MUDMAN:\\Meta-Unlearning with Disruption Masking And Normalization}

\workshoptitle{AI That Keeps Up: Workshop on Continual and Compatible Foundation Model Updates (CCFM)}

\author{Filip Sondej\thanks{Correspondence to: \href{mailto:filip.science921@passinbox.com}{filip.science921@passinbox.com}}\\
  Jagiellonian University \\\And
  Yushi Yang\\
  University of Oxford \\\And
  Mikołaj Kniejski\\
  University of Warsaw\\\And
  Marcel Windys\\
  Independent\\
}

\begin{document}

\maketitle

\begin{abstract}
Language models can retain dangerous knowledge and skills even after extensive safety fine-tuning, posing both misuse and misalignment risks. 
Recent studies show that even specialized unlearning methods can be easily reversed. 
To address this, we systematically evaluate existing ones and propose novel components of unlearning methods and identify ones crucial for irreversible unlearning.

We introduce Disruption Masking, a technique in which we only allow updating weights, where the signs of the unlearning gradient and the retaining gradient are the same. This ensures all updates are non-disruptive.

Additionally, we identify the need for normalizing the unlearning gradients, and also confirm the usefulness of meta-learning. 
We combine these insights into MUDMAN (Meta-Unlearning with Disruption Masking and Normalization) and validate its effectiveness at preventing the recovery of dangerous capabilities. 
MUDMAN outperforms the prior TAR method by 40\%, setting a new state-of-the-art for robust unlearning.
\noindent\textbf{Code:} \href{https://github.com/filyp/MUDMAN}{github.com/filyp/MUDMAN}
\end{abstract}

\section{Introduction}

Language models can acquire dangerous skills during pre-training, such as manipulation, hacking and even knowledge useful for creating bioweapons \citep{li_wmdp_2024}. They may also learn about the safeguards used to control them, which in the future may enable them to subvert such safeguards \citep{greenblatt_alignment_2024,roger_case_2024}.

Studies show that popular safety fine-tuning techniques like DPO and RLHF fail to remove dangerous knowledge; instead, they minimally modify model weights to hide it \citep{lee_mechanistic_2024}, allowing it to reemerge through jailbreak inputs \citep{zou_universal_2023}, or even accidentally \citep{qi_fine-tuning_2023,deeb_unlearning_2024}. Even specialized unlearning techniques turn out to be easily reversible \citep{lynch_eight_2024,lucki_adversarial_2025,deeb_unlearning_2024}.
It remains unclear which components of unlearning algorithms can make harmful capabilities truly irreversible.

To address these challenges, we systematically investigate which components of unlearning algorithms make behavior removal truly irreversible, testing both existing and proposing newly designed ones.
We identify several key components for more robust unlearning: Disruption Masking, meta-learning, and gradient normalization. 
We integrate these into MUDMAN (Meta-Unlearning with Disruption Masking and Normalization) and demonstrate that it significantly outperforms state-of-the-art across multiple models and tasks.

\begin{figure*}[ht]
\centering
\includegraphics[width=1\textwidth,trim={0 7cm 0 7cm},clip]{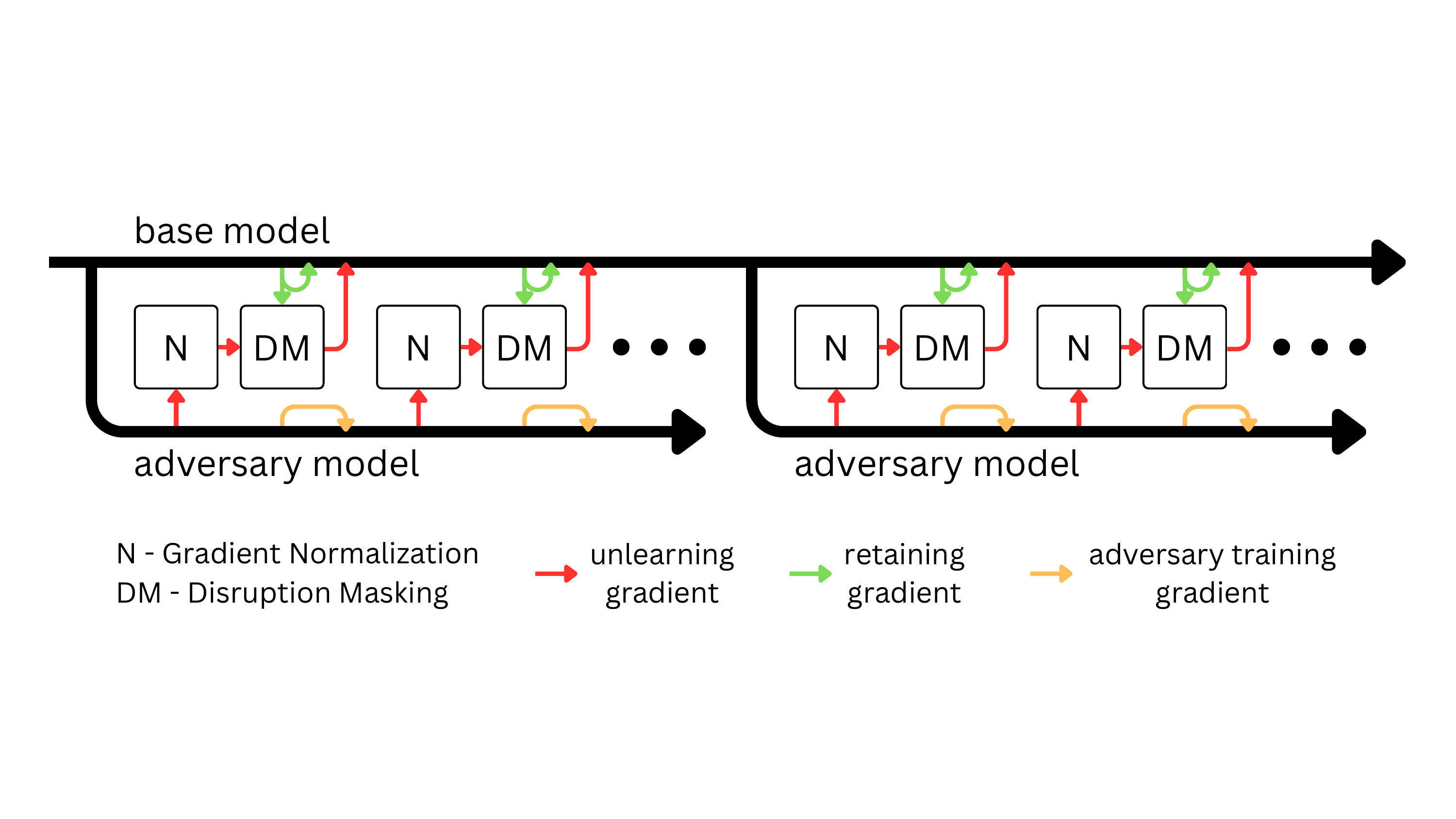}
\caption{\textbf{The MUDMAN pipeline.} 
When training the base model, we periodically fork out an adversary model and train it to perform well on the forget set (orange gradients). Using this adversarial model, we calculate the unlearning gradients (red) to be applied to the base model. Before applying the unlearning gradients, we normalize them and perform Disruption Masking, which zeroes out each weight's unlearning gradient if its sign differs from its retaining gradient (green).}
\label{fig:main}
\end{figure*}


\begin{figure}[ht]
    \centering
    \begin{tabular}{cccc}
        \multicolumn{2}{c}{\textbf{Normal}} &
        \multicolumn{2}{c}{\textbf{Disruption Masking}} \\[0.5em]
        \includegraphics[width=0.21\columnwidth]{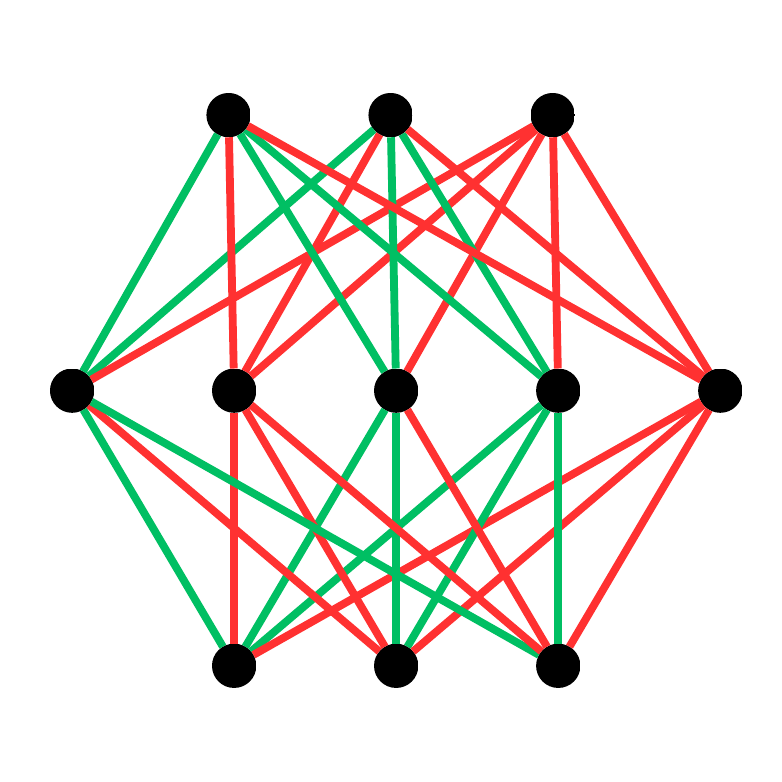} &
        \includegraphics[width=0.21\columnwidth]{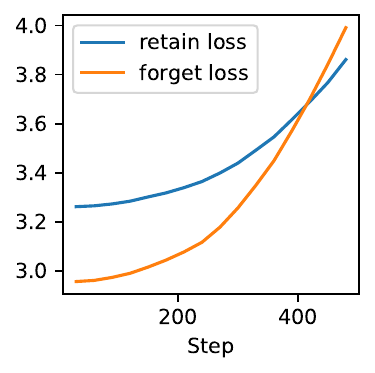} &
        \includegraphics[width=0.21\columnwidth]{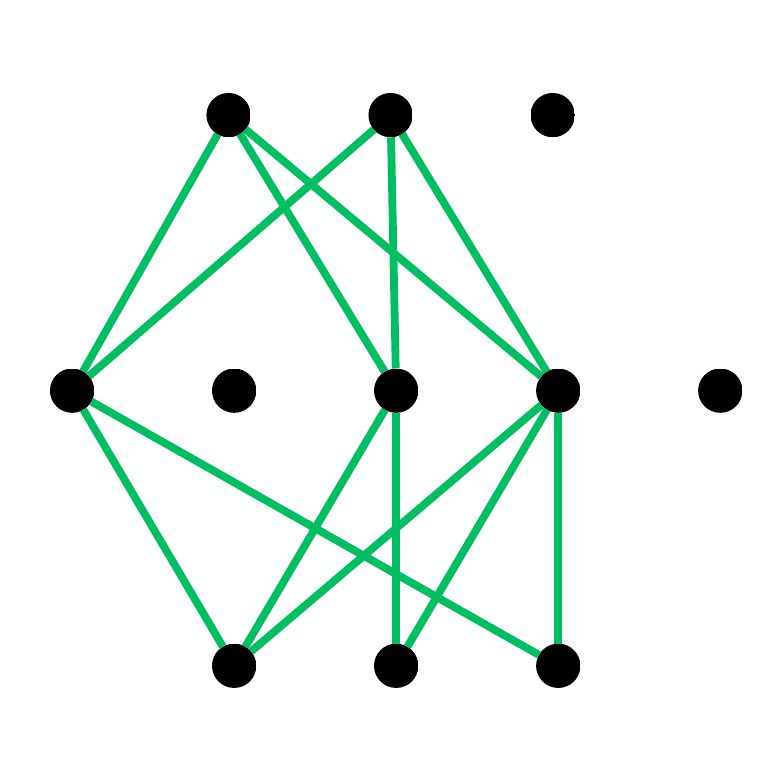} &
        \includegraphics[width=0.21\columnwidth]{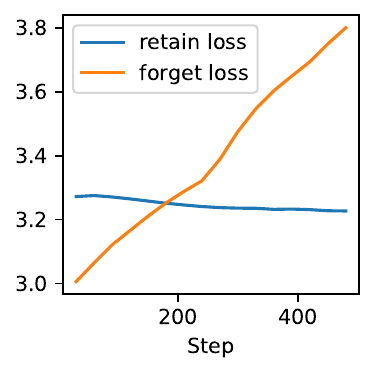}
    \end{tabular}
    \caption{\textbf{Disruption Masking and its training dynamics.}
    \textit{Left}: regular unlearning runs where we maximize the forget loss while trying to minimize the retain loss. For each weight, we color its unlearning update green if it improves the retain loss, and red if it harms it. \textit{Right}: Disruption Masking filters out all these harmful updates. This manages to raise the forget loss without impacting the retain loss.}
    \label{fig:main_diagram}
\end{figure}

\section{Related work}


\paragraph{Current unlearning methods fail to robustly unlearn knowledge} 
Some unlearning approaches disrupt intermediate activations within the model \citep{zou_improving_2024,li_wmdp_2024,rosati_representation_2024}, while others attempt to locate and ablate weights responsible for unwanted behavior \cite{wang_large_2024,wu_depn_2023,uppaal_detox_2024,suau_whispering_2024}. However, \citet{lo_large_2024} found that even when unwanted concepts are directly removed, the model can quickly learn to represent them again using neurons with similar meaning.

To address this, modern unlearning techniques increasingly incorporate MAML (model-agnostic meta-learning) \citep{tamirisa_tamper-resistant_2024,henderson_self-destructing_2023,tamirisa_toward_2024}. This approach anticipates how an attacker could relearn the target capability, by deriving unlearning gradients from a copy of the model trained on the forget set \citep{finn_model-agnostic_2017}.

However, for each existing unlearning methods, there are still ways to elicit the supposedly removed capabilities, such as jailbreak prompts, few-shot prompting, fine-tuning, in-context learning, out-of-distribution inputs or disabling refusal mechanisms with representation engineering \citep{lynch_eight_2024,lucki_adversarial_2025}.

\section{Experiment setup}
\label{sec:methodology}

\paragraph{Models} We conduct experiments with three language models of different sizes: \textbf{pythia-14m} \citep{biderman_pythia_2023}.  \textbf{SmolLM-135M} \citep{allal2024SmolLM} and \textbf{Llama-3.2-1B} \citep{grattafiori2024llama}.
Using pythia-14m allowed us to iterate fast on our techniques, then we validate our findings on the larger, more modern models.

\paragraph{Unlearning datasets} 
We demonstrate \emph{skill unlearning} by attempting to unlearn \textbf{Python} coding ability by using function examples from CodeSearchNet \citep{husain2019codesearchnet}, while retaining the performance on Wikitext \citep{merity2016pointer}. To ensure we do not unintentionally unlearn English, we remove comments and docstrings from the Python dataset. While programming is not a harmful behavior, using loss on the Python dataset provides a high signal-to-noise ratio for comparing unlearning methods and serves as a good baseline.

Then we move to \emph{knowledge unlearning} with a realistic harmful target: unlearning biohazardous knowledge using \textbf{Pile-Bio} \citep{tamirisa_tamper-resistant_2024}, a subset of the Pile \citep{gao_pile_2020} containing texts about molecular biology. 
The rest of the Pile serves as the retain set.

To test unlearning effectiveness on Pile-Bio, we measure the accuracy on \textbf{WMDP-Bio} \citep{li_wmdp_2024}, a dataset of 1273 multiple-choice questions designed as a proxy measurement of hazardous biosecurity knowledge. 
We use temperature=0 to generate the answers.

\paragraph{Hyperparameter search} 
Method performance is highly dependent on hyperparameter selection. To ensure a fair comparison, we perform an automated hyperparameter search for each tested method using Optuna \citep{akiba_optuna_2019}. We verify that each search converges and that the chosen hyperparameter ranges are not saturated. Each search includes hundreds of trials, with each trial consisting of an unlearning stage followed by a fixed relearning stage, and the amount of compute in each stage is constant. (See Appendix~\ref{sec:hyperparameter_searches} for details.)

\paragraph{Unlearning and retaining metrics} 
Optuna tries to maximize forget set loss (or WMDP accuracy) \emph{after relearning} with supervised fine-tuning. 
While we are interested in the overall recoverability of unlearned behavior (e.g., through jailbreaks or out-of-distribution attacks), here we focus on supervised fine-tuning, as it is the most straightforward and most reliable way to resurface removed behavior \citep{lynch_eight_2024}.

In addition to maximizing forget set loss, we ensure that unlearning does not significantly degrade performance on the retain set. To control for this, we terminate and reject trials where the retain loss exceeds a fixed threshold.\footnote{Set as the initial retain loss + 0.05. We add 0.05 to accommodate for random loss fluctuations.}

For the WMDP experiments, rather than controlling for the retain loss increase, we use the accuracy on the MMLU benchmark \citep{hendrycks_measuring_2021} to gain a comprehensive understanding of performance preservation.
\footnote{We use temperature=1 to get MMLU answers, to avoid discontinuous answer switching which happens with temperature=0, which makes trial rejection less predictable.} 
During unlearning we allow at most 1 percentage point drop in MMLU accuracy.

\begin{figure*}[ht]
\centering
\includegraphics[width=1\textwidth]{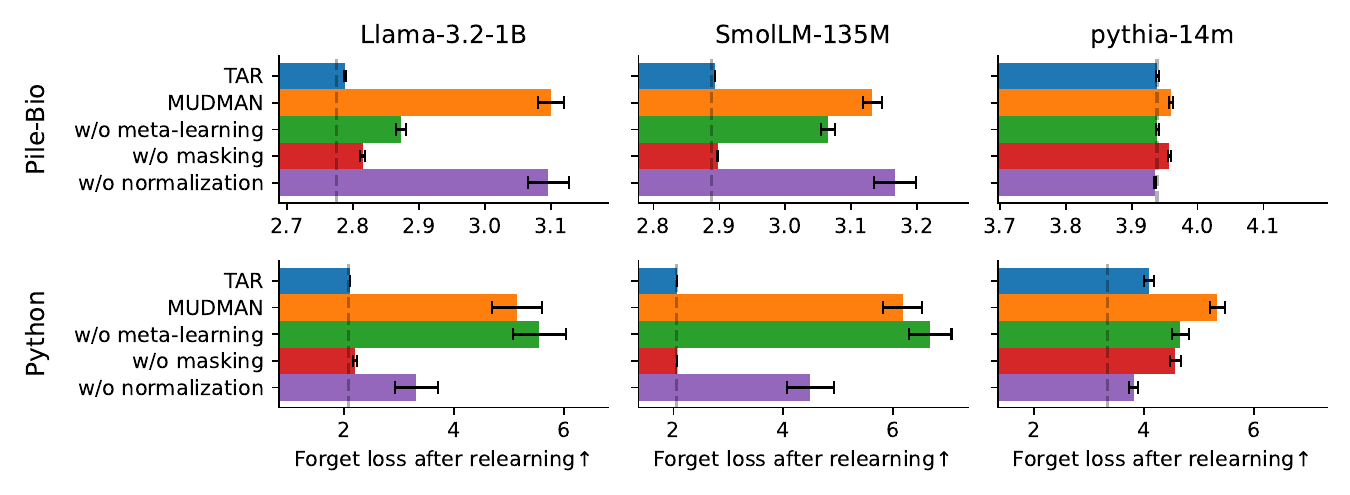}
\caption{\textbf{Ablation study of MUDMAN.} 
To establish that each part of MUDMAN is indeed necessary, we disable them one by one and measure unlearning performance in terms of the forget loss. 
We also compare to the state-of-the-art TAR method. 
The baseline is the loss level with no unlearning applied, but after the same relearning as the other methods underwent. Each bar corresponds to one Optuna hyperparameter search. The reported loss is the average of the last 30 valid trials in that search and error bars are their standard error. 
It shows that Disruption Masking makes a huge difference (orange vs red), and that it accounts for most of the improvement over TAR. Meta-learning and unlearning gradient normalization also help but not in every setup. Sometimes ablating them yields better performance, but insignificantly.
In MUDMAN we use negative cross-entropy as the unlearning loss, due to its top performance.
As discussed in Section~\ref{module_selection}, in each experiment we focus on training the first layers of each MLP.}
\label{fig:ablation_study}
\end{figure*}

\begin{figure}[ht]
\centering
\includegraphics[width=0.5\columnwidth]{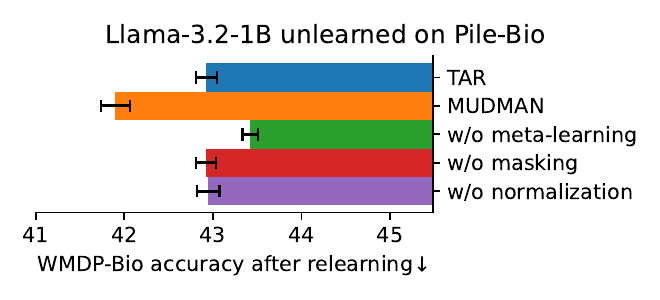}
\caption{\textbf{Accuracy on WMDP-Bio after unlearning and relearning on Pile-Bio.} The base level on the right (45.5\%) is the accuracy with no unlearning applied, but after relearning. 
Reported accuracy is the average of the last 60 valid trials in each hyperparameter search.
We only show Llama, because smaller models had near-random accuracy on WMDP-Bio.
Here, we use MUDMAN with negative entropy \citep{tamirisa_tamper-resistant_2024} as the unlearning loss.
Similarly to Figure~\ref{fig:ablation_study}, using Disruption Masking is crucial (orange vs red) and it accounts for most of the improvement over TAR. There is also clear improvements from meta-learning and unlearning gradient normalization. 
}
\label{fig:wmdp_accuracy}
\end{figure}

\begin{algorithm*}[t]
\caption{MUDMAN -- Meta-Unlearning with Disruption Masking And Normalization}
\textbf{Input:} Model weights $\textcolor{blue}{model}$; retain set $\mathcal{D}_{retain}$; forget set $\mathcal{D}_{forget}$; retain momentum $\mu$; unlearning loss $\mathcal{L}_{unlearning}$; learning rates $\alpha_{retaining}$, $\alpha_{unlearning}$ and $\alpha_{adv}.$
\begin{algorithmic}[1] 
\label{alg:mudman}
\STATE $\textcolor[rgb]{0, 0.392, 0}{retain\_acc} = 0$
  \hfill \text{Initialize retain accumulator}
\FOR{$loop\_num = 1$ \textbf{to} $num\_iterations$}
  \IF{$loop\_num, \text{ is divisible by } fork\_every\_n\_loops$}
      \STATE $\textcolor{orange}{adv} = \textcolor{blue}{model}$
      \hfill \text{Fork out the adversarial model from the main model}
  \ENDIF
  \STATE $x_{retain} \sim \mathcal{D}_{retain}, x_{forget} \sim \mathcal{D}_{forget}$
    \hfill \text{Sample batches}
  \STATE
  \STATE $retain\_grad = \nabla_{\textcolor{blue}{model}} \mathcal{L}_{LM}(\textcolor{blue}{model}, x_{retain})$
    \hfill \text{Calculate retain gradients}
  \STATE $\textcolor[rgb]{0, 0.392, 0}{retain\_acc} = \mu \cdot \textcolor[rgb]{0, 0.392, 0}{retain\_acc} + (1 - \mu) \cdot retain\_grad $
    \hfill \text{Update retain accumulator}
  \STATE $\textcolor{blue}{model} \mathrel{-}= \alpha_{retaining} \textcolor[rgb]{0, 0.392, 0}{retain\_acc}$
    \hfill \text{Update the model}

  \STATE
  \STATE $\textcolor{orange}{adv} \mathrel{-}= \alpha_{adv} \nabla_{\textcolor{orange}{adv}} \mathcal{L}_{LM}(\textcolor{orange}{adv}, x_{forget})$
    \hfill \text{Train adversary on forget set}

  \STATE $\textcolor{red}{grad} = \nabla_{\textcolor{orange}{adv}} \mathcal{L}_{unlearning}(\textcolor{orange}{adv}, x_{forget})$
    \hfill \text{Calculate unlearning gradient}

  \STATE $\textcolor{red}{grad} \mathrel{/}= \| \textcolor{red}{grad} \|_2$
    \hfill \text{Normalize it}
  \STATE $\textcolor{red}{grad} \mathrel{*}= (\operatorname{sign}(\textcolor{red}{grad}) == \operatorname{sign}(\textcolor[rgb]{0, 0.392, 0}{retain\_acc}))$
    \hfill \text{Mask out gradients which hurt retain loss}
  \STATE $\textcolor{blue}{model} \mathrel{-}= \alpha_{unlearning} \textcolor{red}{grad}$
    \hfill \text{Update the model with the unlearning gradient}
\ENDFOR
\end{algorithmic}
\end{algorithm*}

\section{Building a robust unlearning pipeline}
We conducted hundreds of small-scale experiments, testing methods ranging from direct model edits to blocking the updates during relearning.
See Appendix~\ref{sec:failed_methods} for details of approaches.
While almost all methods succeeded in making the forget set loss high during unlearning, relearning typically restored it immediately.

Among them, we identify several key components that consistently improve unlearning robustness. 
We also propose a novel method called Disruption Masking. We integrate these into MUDMAN (Meta-Unlearning with Disruption Masking and Normalization), which outperforms the state-of-the-art TAR method \cite{tamirisa_tamper-resistant_2024}. See Algorithm~\ref{alg:mudman} for pseudocode and Appendix~\ref{sec:mudman_implementation} for a PyTorch implementation. Below, we describe each component of MUDMAN and its overall results.

\subsection{Meta-learning is effective}
First, we confirm the effectiveness of MAML (model-agnostic meta-learning) \citep{finn_model-agnostic_2017}, which is increasingly used in modern unlearning methods \citep{tamirisa_tamper-resistant_2024,henderson_self-destructing_2023,tamirisa_toward_2024}. MAML involves training a copy of the main model, which we call \emph{the adversary}, on the forget set and applying its gradients to the main model.

Despite its adoption, no mechanistic explanation for why MAML helps with robust unlearning has been proposed. We hypothesize that while naive unlearning merely makes harmful circuits dormant \citep{lee_mechanistic_2024} (and dormant circuits cannot be further unlearned because backpropagation cannot locate them), MAML trains an adversary in which these \emph{dormant circuits are reactivated, so that backpropagation can continue to erase them}.

Traditional MAML employs an inner loop to train multiple adversaries and accumulate unlearning gradients before updating the main model \citep{finn_model-agnostic_2017}. However, in early experiments, we found that interleaving adversary and main model updates actually improves performance. 
Therefore, we flatten the process into a single loop (see Algorithm~\ref{alg:mudman}) and focus on training a single adversary deeply, rather than multiple ones briefly.

\subsection{Disruption Masking drives performance}
This is our key contribution from MUDMAN. 
In existing unlearning techniques \citep{tamirisa_tamper-resistant_2024,rosati_representation_2024,li_wmdp_2024}, the model is also trained on the retain set during unlearning, hoping to revert any unintended disruptions. Instead, we adopt a more selective approach, aiming to avoid disruptions altogether rather than correcting them later.

The intuition for such high selectivity is that since the model has already undergone extensive pre-training, its weights are near-optimal. Any unnecessary modifications risk disrupting well-learned representations, requiring significant compute to recover what pre-training had already established. 

We formalise disruption masking as follows: we apply unlearning gradients \textbf{only when they have the same sign as their respective retain gradients} (see Equation~\ref{eq:masking} and Figure~\ref{fig:main_diagram}). This not only prevents disruption but also passively improves the retain performance. 
\begin{equation}
\label{eq:masking}
g_{final} =
    \begin{cases}
        g_{u} & \text{if } \text{sign}(g_{u}) = \text{sign}(g_{r}) \\
        0 & \text{otherwise}
    \end{cases}
\quad
\begin{aligned}
&\text{where } g_{u} \text{ is the unlearning gradient} \\
&\text{and } g_{r} \text{ is the retaining gradient}
\end{aligned}
\end{equation}

Implementation-wise, to generalize beyond the current batch and identify what might break retain set performance overall, we use SGD momentum \citep{rumelhart_learning_1986}, which adds retain gradients to a decaying accumulator. 
Then, instead of the signs of the single-batch retain gradients, we use the signs of this accumulator.
This incurs no additional memory cost, as we already need to store retain gradients alongside unlearning gradients, so we simply store the accumulator instead.



\subsection{Gradient normalization improves speed and stability}

In the later phase of the unlearning process, 
we observe that gradient norms shrink significantly, causing unlearning to slow down. To address this, we normalize the unlearning gradients so that the strength of unlearning remains constant. 
Specifically, we divide each unlearning gradient by a global gradient norm, calculated as:
$(\sum_{m \in \text{modules}} \|unlearning\_grad_m\|^2)^{0.5}$.

This not only improves overall performance but also makes the training process easier to tune. See Appendix~\ref{sec:gradient_normalization} for a comparison of other normalization variants.

\subsection{Module selection improves selectivity}
\label{module_selection}
We selectively filter out model modules and test which ones to intervene on for more selective unlearning.
On Figure~\ref{fig:target_modules} we see that MLP's second weight matrices and attention's Q and K matrices tend to significantly disrupt general performance, while not being very important for unlearning performance.
The most effective unlearning targets are the first layers of MLPs, and in the case of gated MLPs, the gating matrices.
This is likely because only these modules are able to deactivate MLP neurons, and once a neuron is inactive, neither its incoming nor outgoing weights can be updated by backpropagation, effectively preventing relearning.
See Appendix~\ref{sec:target_modules} for detailed comparisons of unlearning on various modules.

\subsection{Improvements from MUDMAN}
We validate the superiority of MUDMAN on three models and two unlearning tasks over TAR \citep{tamirisa_tamper-resistant_2024}, a state-of-the-art unlearning method. 
For a clean comparison, we adapt the TAR implementation to match our setup: using a single training loop and omitting its initial representation noising step.
As shown in Figure~\ref{fig:ablation_study}, \emph{MUDMAN consistently outperforms this adapted TAR} across all tested cases.
In individual setups, while some ablations occasionally match the full MUDMAN, none of them ever significantly surpasses it.


\section{Conclusion}

We introduce MUDMAN, a novel unlearning pipeline consisting of three components: meta-unlearning, Disruption Masking and normalization of unlearning gradients. 
We show across multiple datasets and models that each of these components significantly improves the robustness of unlearning of harmful knowledge. 
They also come at minimal computational and memory overhead and are compatible with any unlearning loss.

Our results demonstrate the success of selective unlearning methods which do not disrupt model performance in the first place, inspiring future work to pursue this line of research. Our findings bring closer to the goal of truly irreversible unlearning, which is critical for safe deployment as language models as they continue to acquire dangerous knowledge and capabilities.

\section{Limitations}

\paragraph{More elicitation methods}
In this study, we only use supervised fine-tuning to elicit unwanted behavior. 
While it is the most powerful and reliable elicitation method \citep{lynch_eight_2024}, future work could extend to evaluate other attacks, in which the attacker does not have weight access, for example with jailbreak prompts \citep{zou_universal_2023}.

To further test whether our techniques truly remove capabilities or just make them harder to recover, one could use the approach introduced by \citet{deeb_unlearning_2024}, where the attacker tries to uncover some unknown, supposedly unlearned facts, by fine-tuning on another, non-overlapping set of facts.


\paragraph{Stacking with other methods} For simplicity, we only evaluate monolithic methods which apply a single algorithm throughout the whole unlearning process. 
This involved stripping our TAR baseline \citep{tamirisa_tamper-resistant_2024} of its initial representation noising step and only keeping the meta-learning core of the algorithm. Future work could investigate the effectiveness of consecutively stacking different unlearning methods, looking for synergies between them.

\paragraph{More work on selectivity and granularity}
In our experiments, we generally see good results from techniques aiming to be more selective and granular. 
Future work could try to understand what happens at the level of individual logits and tweaking unlearning loss functions to target only the crucial logits. Additionally, a more granular, per-token analysis of forget set loss could be valuable: rather than concentrating the loss increase on a few unwanted tokens, the goal could be set to a high loss across all unwanted tokens.


\begin{ack}

We thank Fabien Roger, Stephen Casper, Adam Mahdi, Kay Kozaronek and Artyom Karpov for valuable discussions and feedback.

Filip Sondej's work was partially funded by a grant from Coefficient Giving.

\end{ack}

\bibliographystyle{plainnat}
\bibliography{from_yushi_dpo_paper, from_filip_zotero}

@misc{merity2016pointer,
      title={Pointer Sentinel Mixture Models},
      author={Stephen Merity and Caiming Xiong and James Bradbury and Richard Socher},
      year={2016},
      eprint={1609.07843},
      archivePrefix={arXiv},
      primaryClass={cs.CL}
}

@article{rumelhart_learning_1986,
	title = {Learning representations by back-propagating errors},
	volume = {323},
	issn = {0028-0836},
	url = {https://ui.adsabs.harvard.edu/abs/1986Natur.323..533R},
	doi = {10.1038/323533a0},
	urldate = {2025-02-14},
	journal = {Nature},
	author = {Rumelhart, David E. and Hinton, Geoffrey E. and Williams, Ronald J.},
	month = oct,
	year = {1986},
	note = {ADS Bibcode: 1986Natur.323..533R},
	pages = {533--536},
}

@misc{zou_universal_2023,
	title = {Universal and {Transferable} {Adversarial} {Attacks} on {Aligned} {Language} {Models}},
	url = {http://arxiv.org/abs/2307.15043},
	doi = {10.48550/arXiv.2307.15043},
	urldate = {2025-02-14},
	publisher = {arXiv},
	author = {Zou, Andy and Wang, Zifan and Carlini, Nicholas and Nasr, Milad and Kolter, J. Zico and Fredrikson, Matt},
	month = dec,
	year = {2023},
	note = {arXiv:2307.15043 [cs]},
}

@misc{grattafiori2024llama,
      title={The Llama 3 Herd of Models}, 
      author={Aaron Grattafiori and others},
      year={2024},
      eprint={2407.21783},
      archivePrefix={arXiv},
      primaryClass={cs.AI}
}

@misc{schwarz_progress_2018,
	title = {Progress \& {Compress}: {A} scalable framework for continual learning},
	shorttitle = {Progress \& {Compress}},
	url = {http://arxiv.org/abs/1805.06370},
	doi = {10.48550/arXiv.1805.06370},
	urldate = {2025-02-14},
	publisher = {arXiv},
	author = {Schwarz, Jonathan and Luketina, Jelena and Czarnecki, Wojciech M. and Grabska-Barwinska, Agnieszka and Teh, Yee Whye and Pascanu, Razvan and Hadsell, Raia},
	month = jul,
	year = {2018},
	note = {arXiv:1805.06370 [stat]},
}

@misc{chaudhry_efficient_2019,
	title = {Efficient {Lifelong} {Learning} with {A}-{GEM}},
	url = {http://arxiv.org/abs/1812.00420},
	doi = {10.48550/arXiv.1812.00420},
	urldate = {2025-02-14},
	publisher = {arXiv},
	author = {Chaudhry, Arslan and Ranzato, Marc'Aurelio and Rohrbach, Marcus and Elhoseiny, Mohamed},
	month = jan,
	year = {2019},
	note = {arXiv:1812.00420 [cs]},
}

@misc{lucki_adversarial_2025,
	title = {An {Adversarial} {Perspective} on {Machine} {Unlearning} for {AI} {Safety}},
	url = {http://arxiv.org/abs/2409.18025},
	doi = {10.48550/arXiv.2409.18025},
	urldate = {2025-02-12},
	publisher = {arXiv},
	author = {Łucki, Jakub and Wei, Boyi and Huang, Yangsibo and Henderson, Peter and Tramèr, Florian and Rando, Javier},
	month = jan,
	year = {2025},
	note = {arXiv:2409.18025 [cs]},
}

@misc{greenblatt_alignment_2024,
	title = {Alignment faking in large language models},
	url = {http://arxiv.org/abs/2412.14093},
	doi = {10.48550/arXiv.2412.14093},
	urldate = {2025-02-12},
	publisher = {arXiv},
	author = {Greenblatt, Ryan and Denison, Carson and Wright, Benjamin and Roger, Fabien and MacDiarmid, Monte and Marks, Sam and Treutlein, Johannes and Belonax, Tim and Chen, Jack and Duvenaud, David and Khan, Akbir and Michael, Julian and Mindermann, Sören and Perez, Ethan and Petrini, Linda and Uesato, Jonathan and Kaplan, Jared and Shlegeris, Buck and Bowman, Samuel R. and Hubinger, Evan},
	month = dec,
	year = {2024},
	note = {arXiv:2412.14093 [cs]},
}

@misc{hendrycks_measuring_2021,
	title = {Measuring {Massive} {Multitask} {Language} {Understanding}},
	url = {http://arxiv.org/abs/2009.03300},
	doi = {10.48550/arXiv.2009.03300},
	urldate = {2025-02-12},
	publisher = {arXiv},
	author = {Hendrycks, Dan and Burns, Collin and Basart, Steven and Zou, Andy and Mazeika, Mantas and Song, Dawn and Steinhardt, Jacob},
	month = jan,
	year = {2021},
	note = {arXiv:2009.03300 [cs]},
}

@misc{guan_deliberative_2025,
	title = {Deliberative {Alignment}: {Reasoning} {Enables} {Safer} {Language} {Models}},
	shorttitle = {Deliberative {Alignment}},
	url = {http://arxiv.org/abs/2412.16339},
	doi = {10.48550/arXiv.2412.16339},
	urldate = {2025-02-09},
	publisher = {arXiv},
	author = {Guan, Melody Y. and Joglekar, Manas and Wallace, Eric and Jain, Saachi and Barak, Boaz and Helyar, Alec and Dias, Rachel and Vallone, Andrea and Ren, Hongyu and Wei, Jason and Chung, Hyung Won and Toyer, Sam and Heidecke, Johannes and Beutel, Alex and Glaese, Amelia},
	month = jan,
	year = {2025},
	note = {arXiv:2412.16339 [cs]},
}

@article{roger_case_2024,
	title = {The case for unlearning that removes information from {LLM} weights},
	url = {https://www.lesswrong.com/posts/9AbYkAy8s9LvB7dT5/the-case-for-unlearning-that-removes-information-from-llm},
	language = {en},
	urldate = {2025-02-09},
	author = {Roger, Fabien},
	month = oct,
	year = {2024},
}

@article{husain2019codesearchnet,
  title={{CodeSearchNet} challenge: Evaluating the state of semantic code search},
  author={Husain, Hamel and Wu, Ho-Hsiang and Gazit, Tiferet and Allamanis, Miltiadis and Brockschmidt, Marc},
  journal={arXiv preprint arXiv:1909.09436},
  year={2019}
}

@misc{biderman_pythia_2023,
	title = {Pythia: {A} {Suite} for {Analyzing} {Large} {Language} {Models} {Across} {Training} and {Scaling}},
	shorttitle = {Pythia},
	url = {http://arxiv.org/abs/2304.01373},
	doi = {10.48550/arXiv.2304.01373},
	urldate = {2025-02-08},
	publisher = {arXiv},
	author = {Biderman, Stella and Schoelkopf, Hailey and Anthony, Quentin and Bradley, Herbie and O'Brien, Kyle and Hallahan, Eric and Khan, Mohammad Aflah and Purohit, Shivanshu and Prashanth, USVSN Sai and Raff, Edward and Skowron, Aviya and Sutawika, Lintang and Wal, Oskar van der},
	month = may,
	year = {2023},
	note = {arXiv:2304.01373 [cs]},
}

@misc{allal2024SmolLM,
      title={SmolLM - blazingly fast and remarkably powerful}, 
      author={Loubna Ben Allal and Anton Lozhkov and Elie Bakouch and Leandro von Werra and Thomas Wolf},
      year={2024},
}

@misc{akiba_optuna_2019,
	title = {Optuna: {A} {Next}-generation {Hyperparameter} {Optimization} {Framework}},
	shorttitle = {Optuna},
	url = {http://arxiv.org/abs/1907.10902},
	doi = {10.48550/arXiv.1907.10902},
	urldate = {2025-02-04},
	publisher = {arXiv},
	author = {Akiba, Takuya and Sano, Shotaro and Yanase, Toshihiko and Ohta, Takeru and Koyama, Masanori},
	month = jul,
	year = {2019},
	note = {arXiv:1907.10902 [cs]},
}

@misc{gao_pile_2020,
	title = {The {Pile}: {An} {800GB} {Dataset} of {Diverse} {Text} for {Language} {Modeling}},
	shorttitle = {The {Pile}},
	url = {http://arxiv.org/abs/2101.00027},
	doi = {10.48550/arXiv.2101.00027},
	urldate = {2025-02-04},
	publisher = {arXiv},
	author = {Gao, Leo and Biderman, Stella and Black, Sid and Golding, Laurence and Hoppe, Travis and Foster, Charles and Phang, Jason and He, Horace and Thite, Anish and Nabeshima, Noa and Presser, Shawn and Leahy, Connor},
	month = dec,
	year = {2020},
	note = {arXiv:2101.00027 [cs]},
}

@misc{finn_model-agnostic_2017,
	title = {Model-{Agnostic} {Meta}-{Learning} for {Fast} {Adaptation} of {Deep} {Networks}},
	url = {http://arxiv.org/abs/1703.03400},
	doi = {10.48550/arXiv.1703.03400},
	urldate = {2024-09-04},
	publisher = {arXiv},
	author = {Finn, Chelsea and Abbeel, Pieter and Levine, Sergey},
	month = jul,
	year = {2017},
	note = {arXiv:1703.03400 [cs]},
}

@article{tamirisa_toward_2024,
	title = {{TOWARD} {ROBUST} {UNLEARNING} {FOR} {LLMS}},
	language = {en},
	author = {Tamirisa, Rishub and Bharathi, Bhrugu and Zhou, Andy and Li, Bo and Mazeika, Mantas},
	year = {2024},
}

@misc{rosati_representation_2024,
	title = {Representation noising effectively prevents harmful fine-tuning on {LLMs}},
	url = {http://arxiv.org/abs/2405.14577},
	doi = {10.48550/arXiv.2405.14577},
	urldate = {2024-07-26},
	publisher = {arXiv},
	author = {Rosati, Domenic and Wehner, Jan and Williams, Kai and Bartoszcze, Łukasz and Atanasov, David and Gonzales, Robie and Majumdar, Subhabrata and Maple, Carsten and Sajjad, Hassan and Rudzicz, Frank},
	month = may,
	year = {2024},
	note = {arXiv:2405.14577 [cs]},
}

@misc{li_wmdp_2024,
	title = {The {WMDP} {Benchmark}: {Measuring} and {Reducing} {Malicious} {Use} {With} {Unlearning}},
	shorttitle = {The {WMDP} {Benchmark}},
	url = {http://arxiv.org/abs/2403.03218},
	doi = {10.48550/arXiv.2403.03218},
	urldate = {2024-07-26},
	publisher = {arXiv},
	author = {Li, Nathaniel and Pan, Alexander and Gopal, Anjali and Yue, Summer and Berrios, Daniel and Gatti, Alice and Li, Justin D. and Dombrowski, Ann-Kathrin and Goel, Shashwat and Phan, Long and Mukobi, Gabriel and Helm-Burger, Nathan and Lababidi, Rassin and Justen, Lennart and Liu, Andrew B. and Chen, Michael and Barrass, Isabelle and Zhang, Oliver and Zhu, Xiaoyuan and Tamirisa, Rishub and Bharathi, Bhrugu and Khoja, Adam and Zhao, Zhenqi and Herbert-Voss, Ariel and Breuer, Cort B. and Marks, Samuel and Patel, Oam and Zou, Andy and Mazeika, Mantas and Wang, Zifan and Oswal, Palash and Lin, Weiran and Hunt, Adam A. and Tienken-Harder, Justin and Shih, Kevin Y. and Talley, Kemper and Guan, John and Kaplan, Russell and Steneker, Ian and Campbell, David and Jokubaitis, Brad and Levinson, Alex and Wang, Jean and Qian, William and Karmakar, Kallol Krishna and Basart, Steven and Fitz, Stephen and Levine, Mindy and Kumaraguru, Ponnurangam and Tupakula, Uday and Varadharajan, Vijay and Wang, Ruoyu and Shoshitaishvili, Yan and Ba, Jimmy and Esvelt, Kevin M. and Wang, Alexandr and Hendrycks, Dan},
	month = may,
	year = {2024},
	note = {arXiv:2403.03218 [cs]},
}

@misc{lo_large_2024,
	title = {Large {Language} {Models} {Relearn} {Removed} {Concepts}},
	url = {http://arxiv.org/abs/2401.01814},
	language = {en},
	urldate = {2024-07-26},
	publisher = {arXiv},
	author = {Lo, Michelle and Cohen, Shay B. and Barez, Fazl},
	month = jan,
	year = {2024},
	note = {arXiv:2401.01814 [cs]},
}

@misc{henderson_self-destructing_2023,
	title = {Self-{Destructing} {Models}: {Increasing} the {Costs} of {Harmful} {Dual} {Uses} of {Foundation} {Models}},
	shorttitle = {Self-{Destructing} {Models}},
	url = {http://arxiv.org/abs/2211.14946},
	doi = {10.48550/arXiv.2211.14946},
	urldate = {2024-07-26},
	publisher = {arXiv},
	author = {Henderson, Peter and Mitchell, Eric and Manning, Christopher D. and Jurafsky, Dan and Finn, Chelsea},
	month = aug,
	year = {2023},
	note = {arXiv:2211.14946 [cs]},
}

@misc{lee_mechanistic_2024,
	title = {A {Mechanistic} {Understanding} of {Alignment} {Algorithms}: {A} {Case} {Study} on {DPO} and {Toxicity}},
	shorttitle = {A {Mechanistic} {Understanding} of {Alignment} {Algorithms}},
	url = {http://arxiv.org/abs/2401.01967},
	doi = {10.48550/arXiv.2401.01967},
	urldate = {2024-07-26},
	publisher = {arXiv},
	author = {Lee, Andrew and Bai, Xiaoyan and Pres, Itamar and Wattenberg, Martin and Kummerfeld, Jonathan K. and Mihalcea, Rada},
	month = jan,
	year = {2024},
	note = {arXiv:2401.01967 [cs]},
}

@misc{qi_fine-tuning_2023,
	title = {Fine-tuning {Aligned} {Language} {Models} {Compromises} {Safety}, {Even} {When} {Users} {Do} {Not} {Intend} {To}!},
	url = {http://arxiv.org/abs/2310.03693},
	doi = {10.48550/arXiv.2310.03693},
	urldate = {2024-07-29},
	publisher = {arXiv},
	author = {Qi, Xiangyu and Zeng, Yi and Xie, Tinghao and Chen, Pin-Yu and Jia, Ruoxi and Mittal, Prateek and Henderson, Peter},
	month = oct,
	year = {2023},
	note = {arXiv:2310.03693 [cs]},
}

@misc{tamirisa_tamper-resistant_2024,
	title = {Tamper-{Resistant} {Safeguards} for {Open}-{Weight} {LLMs}},
	url = {http://arxiv.org/abs/2408.00761},
	urldate = {2024-08-05},
	publisher = {arXiv},
	author = {Tamirisa, Rishub and Bharathi, Bhrugu and Phan, Long and Zhou, Andy and Gatti, Alice and Suresh, Tarun and Lin, Maxwell and Wang, Justin and Wang, Rowan and Arel, Ron and Zou, Andy and Song, Dawn and Li, Bo and Hendrycks, Dan and Mazeika, Mantas},
	month = aug,
	year = {2024},
	note = {arXiv:2408.00761 [cs]},
}

@misc{zou_improving_2024,
	title = {Improving {Alignment} and {Robustness} with {Circuit} {Breakers}},
	url = {http://arxiv.org/abs/2406.04313},
	doi = {10.48550/arXiv.2406.04313},
	urldate = {2024-08-22},
	publisher = {arXiv},
	author = {Zou, Andy and Phan, Long and Wang, Justin and Duenas, Derek and Lin, Maxwell and Andriushchenko, Maksym and Wang, Rowan and Kolter, Zico and Fredrikson, Matt and Hendrycks, Dan},
	month = jul,
	year = {2024},
	note = {arXiv:2406.04313 [cs]},
}

@misc{wu_depn_2023,
	title = {{DEPN}: {Detecting} and {Editing} {Privacy} {Neurons} in {Pretrained} {Language} {Models}},
	shorttitle = {{DEPN}},
	url = {http://arxiv.org/abs/2310.20138},
	doi = {10.48550/arXiv.2310.20138},
	urldate = {2024-09-05},
	publisher = {arXiv},
	author = {Wu, Xinwei and Li, Junzhuo and Xu, Minghui and Dong, Weilong and Wu, Shuangzhi and Bian, Chao and Xiong, Deyi},
	month = dec,
	year = {2023},
	note = {arXiv:2310.20138 [cs]},
}

@misc{wang_large_2024,
	title = {Large {Scale} {Knowledge} {Washing}},
	url = {https://arxiv.org/abs/2405.16720v2},
	language = {en},
	urldate = {2024-09-05},
	journal = {arXiv.org},
	author = {Wang, Yu and Wu, Ruihan and He, Zexue and Chen, Xiusi and McAuley, Julian},
	month = may,
	year = {2024},
}

@misc{lynch_eight_2024,
	title = {Eight {Methods} to {Evaluate} {Robust} {Unlearning} in {LLMs}},
	url = {http://arxiv.org/abs/2402.16835},
	doi = {10.48550/arXiv.2402.16835},
	urldate = {2024-10-01},
	publisher = {arXiv},
	author = {Lynch, Aengus and Guo, Phillip and Ewart, Aidan and Casper, Stephen and Hadfield-Menell, Dylan},
	month = feb,
	year = {2024},
	note = {arXiv:2402.16835 [cs]},
}

@misc{uppaal_detox_2024,
	title = {{DeTox}: {Toxic} {Subspace} {Projection} for {Model} {Editing}},
	shorttitle = {{DeTox}},
	url = {http://arxiv.org/abs/2405.13967},
	doi = {10.48550/arXiv.2405.13967},
	urldate = {2024-10-09},
	publisher = {arXiv},
	author = {Uppaal, Rheeya and Dey, Apratim and He, Yiting and Zhong, Yiqiao and Hu, Junjie},
	month = may,
	year = {2024},
	note = {arXiv:2405.13967 [cs]},
}

@misc{ji_beavertails_2023,
	title = {{BeaverTails}: {Towards} {Improved} {Safety} {Alignment} of {LLM} via a {Human}-{Preference} {Dataset}},
	shorttitle = {{BeaverTails}},
	url = {http://arxiv.org/abs/2307.04657},
	doi = {10.48550/arXiv.2307.04657},
	urldate = {2024-11-23},
	publisher = {arXiv},
	author = {Ji, Jiaming and Liu, Mickel and Dai, Juntao and Pan, Xuehai and Zhang, Chi and Bian, Ce and Zhang, Chi and Sun, Ruiyang and Wang, Yizhou and Yang, Yaodong},
	month = nov,
	year = {2023},
	note = {arXiv:2307.04657},
}

@misc{deeb_unlearning_2024,
	title = {Do {Unlearning} {Methods} {Remove} {Information} from {Language} {Model} {Weights}?},
	url = {http://arxiv.org/abs/2410.08827},
	doi = {10.48550/arXiv.2410.08827},
	urldate = {2024-11-28},
	publisher = {arXiv},
	author = {Deeb, Aghyad and Roger, Fabien},
	month = nov,
	year = {2024},
	note = {arXiv:2410.08827},
}

@misc{suau_whispering_2024,
	title = {Whispering {Experts}: {Neural} {Interventions} for {Toxicity} {Mitigation} in {Language} {Models}},
	shorttitle = {Whispering {Experts}},
	url = {http://arxiv.org/abs/2407.12824},
	doi = {10.48550/arXiv.2407.12824},
	urldate = {2024-11-28},
	publisher = {arXiv},
	author = {Suau, Xavier and Delobelle, Pieter and Metcalf, Katherine and Joulin, Armand and Apostoloff, Nicholas and Zappella, Luca and Rodríguez, Pau},
	month = jul,
	year = {2024},
	note = {arXiv:2407.12824},
}


\appendix

\section{Future work}

\paragraph{Dataset design} 
The spirit of making unlearning more selective can be also applied to dataset design. Usually, the bulk of the forget set is made of benign tokens (filler words, irrelevant passages, etc.), and we are only interested in breaking a smaller subset of critical tokens. So we should find a scalable way to label the importance of each token in the forget sets we use. This would aid both in unlearning that does not disrupt the model, and in a more accurate evaluation of unlearning methods.

\paragraph{Interpretability} We believe that a lot of insight can be extracted from using interpretability on the effects of unlearning techniques. One thing we find particularly mysterious, is that during relearning we often find that forget loss decrease is stopped at some high level and can stay like this for a long time. Then it sometimes suddenly "breaks through", which is reminiscent of grokking. We would like to understand what exactly happens around this "break through", and what mechanistically differentiates cases where this happens from ones where it does not. Also, if we disallow fine-tuning, are the models before "break through" more resistant to other elicitation techniques? Interpretability could start by identifying which neurons are affected by unlearning the most, how their triggering context is shifted and how they relate to other neurons.

\paragraph{Adversary depth} Unlearning can happen at different "depths", defined as the length of training of the adversary. Unlearning with shallow adversaries or no adversary at all tends to be easily reverted, while more depth can be too computationally costly. It would be illuminating to check in isolation what happens at various depths.


\paragraph{Relation to deliberative alignment} Finally, the concept of unlearning could be inverted, and we could try to make sure that some critical desired behavior \emph{always} happens, and will never be disturbed out-of-distribution or under jailbreaks or fine-tuning. This is particularly useful in the deliberative alignment framework \citep{guan_deliberative_2025}, where we want some lines of reasoning to trigger reliably in certain contexts.


\section{Target modules}
\label{sec:target_modules}

\begin{figure*}[h]
    \centering
    \includegraphics[width=\linewidth]{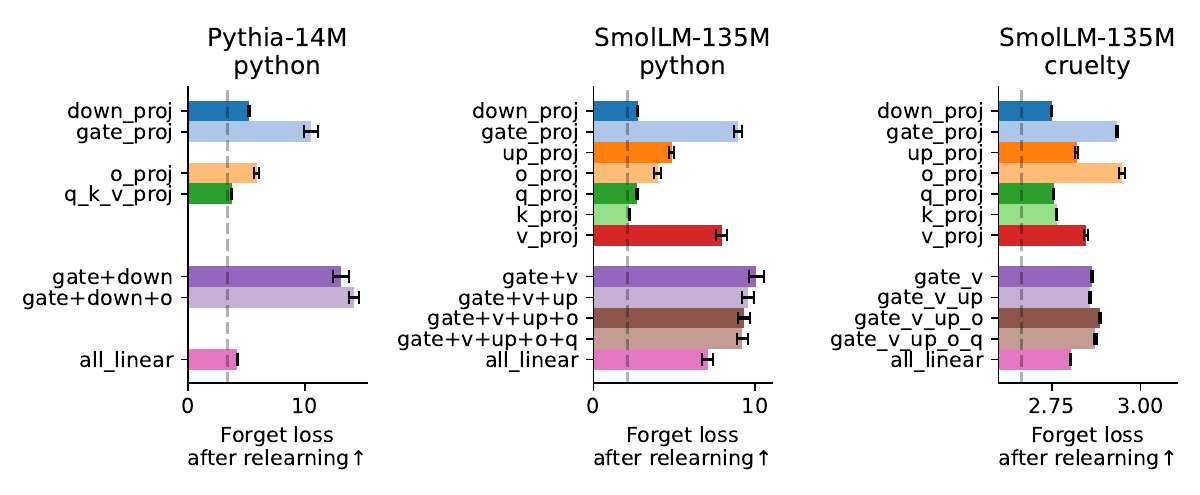}
    \caption{\textbf{Performance comparison across different target module configurations for unlearning.} Higher values indicate better unlearning effectiveness while maintaining model capabilities. The baseline is the loss without any unlearning, but with the same relearning stage as all the methods underwent. Gate projection (and in the case of pythia its equivalent--the first MLP layer) helps most consistently. Other potential candidates for intervention are V, O and up projections. Q, K and down projections disrupt retain performance so much that is it better to omit them. In case of Pythia, Q, K and V matrices are integrated into one module, so we were not able to analyze them in separation.}
    \label{fig:target_modules}
\end{figure*}

We have seen that some modules of the model are not important for unlearning but they disrupt the general performance a lot. It is better to freeze such modules and focus only on the ones where unlearning and retaining performance can be better separated. Freezing modules also lets us save memory, because we do not need to store their gradients, adversarial weights and accumulators of retain gradients. See Figure~\ref{fig:target_modules} for a comparison of unlearning robustness for various modules and their configurations.

Both Llama-3.2-1B and SmolLM-135M use gated MLP architecture, and for them we find unlearning on the gate\_proj module is the most effective. For pythia-14m, which uses traditional MLPs with two layers: up\_proj and down\_proj, intervening on up\_proj performs the best. This is actually consistent--in both cases the optimal module is the one whose outputs are passed to the MLP's activation function. 


\section{Types of gradient normalization}
\label{sec:gradient_normalization}

In Table~\ref{tab:normalization} we compare different ways of normalizing the unlearning gradients. While the difference in performance with the case without any normalization is significant, the differences between various normalization types are less clear.

The naive way is to normalize each module of the model separately. But this may wrongly inflate gradients on some overall low-norm modules. It is safer to normalize using a global gradient norm calculated as $(\sum_{m \in \text{modules}} \|\nabla_m \mathcal{L}_{LM}\|^2)^{0.5}$

We can also normalize before or after applying the gradient masking. Turns out it does not make much difference in performance, so we just go with normalizing before the mask, which has a simpler implementation.

\begin{table}[h]
\caption{Comparison of different gradient normalization strategies.
We report forget loss after 600 unlearning passes using MUDMAN and 300 relearning passes. We let Optuna optimize the hyperparameters for 600 trials and then measure the mean and standard error of the last 50 completed trials. Higher values indicate better performance. (We use the \texttt{pythia-14m} model the \texttt{python} forget set. For more details see \texttt{configs/pythia\_normalization\_test.yaml} in our repository.)}
\centering
\begin{tabular}{cl}
\hline
\textbf{Normalization type} & \textbf{Forget loss ↑} \\
\hline
No Normalization & 4.80±0.14 \\
Per Parameter & \textbf{7.61}±0.48 \\
Global Post Mask & \textbf{7.60}±0.49 \\
\textbf{Global Pre Mask} & \textbf{8.51}±0.55 \\
\hline
\end{tabular}
\label{tab:normalization}
\end{table}

\section{Unlearning tendencies}

We had a partial success unlearning dangerous tendencies. We chose cruelty as our candidate, and as the forget set we used animal abuse examples from BeaverTails \citep{ji_beavertails_2023}. It contains Q\&A pairs, with questions like "What are some ways to torture a small animal?". As the retain set, we used pairs from BeaverTails with benign answers.

\begin{figure*}
    \centering
    \includegraphics[width=\linewidth]{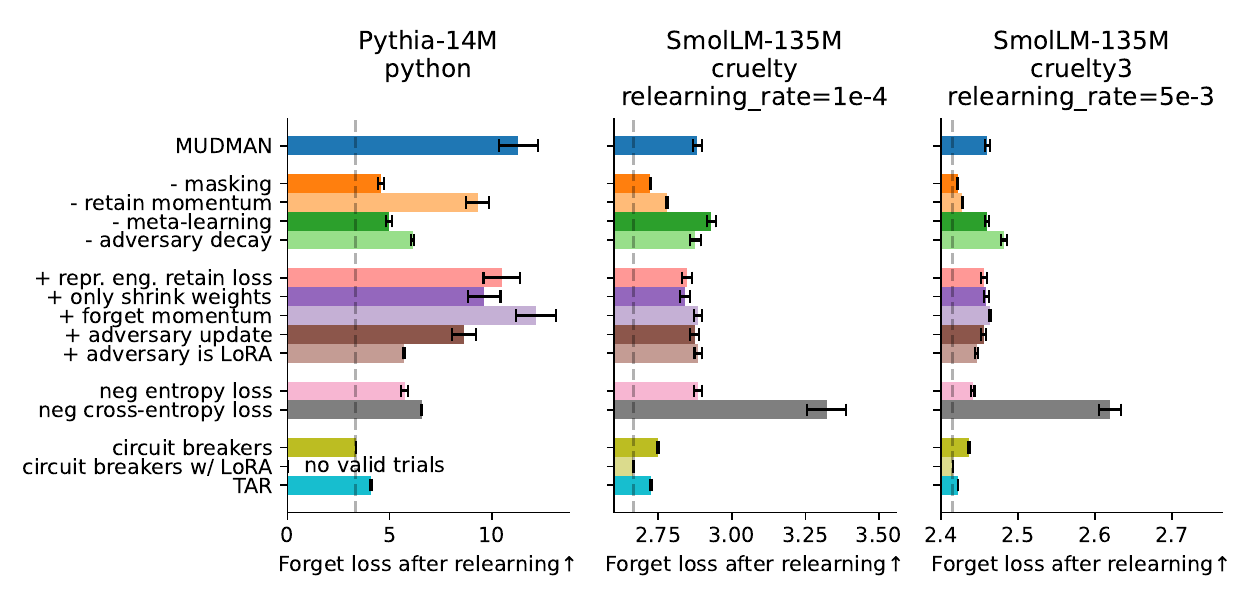}
      \caption{\textbf{Robustness of unlearning cruel tendencies.} Each bar is one Optuna hyperparameter search. The baselines are loss levels with no unlearning applied, but \emph{also after the same relearning} as the other methods underwent. We can see some benefit of using Disruption Masking. Puzzlingly, meta-learning does not help at all, suggesting that cruel tendencies may be more "shallow" than the skills and knowledge we tried to unlearn in other experiments. The biggest effect is caused by simply using cross-entropy loss for unlearning (here in MUDMAN and its variants we used selective logit loss, described in Appendix~\ref{sec:failed_methods}). We conclude that unlearning tendencies like cruelty seems possible, but will require more refinement of the methods. We also show performance for some of rejected methods, that we describe in Appendix~\ref{sec:failed_methods}.}
    \label{fig:failed_methods_and_cruelty}
\end{figure*}

On Figure~\ref{fig:failed_methods_and_cruelty} we show unlearning performance on this cruelty dataset, as well as on Python for comparison. You can also see the performance of some of our failed methods that we describe in Appendix~\ref{sec:failed_methods} below.

\section{Failed methods}
\label{sec:failed_methods}


In our setup with pythia-14m and the Python forget set, we tested dozens of components of unlearning methods, both existing ones and ones designed by us.
In the main text we presented the best performing ones, for which we also conducted experiments on other models and datasets.
Here, we would like to report the negative results, to inform future explorations. 

Note that most of these components have been tested in isolation, so we may have missed some synergies. In particular, many of them have not been used together with meta-learning. We also do not rule out that some of them may prove useful when performance disruption is measured more fully, using capability benchmarks.

Components we tested can be divided into five categories presented in subsections below: dampening relearning gradients, direct weight edits, erasing capabilities rather than drowning them out, making unlearning more selective, and tweaking the meta-learning. The methods were tested in many variants and combinations, but for conciseness we only describe them individually, and omit some details of the variants.

\subsection{Dampening relearning gradients}

Relearning will not happen if the gradients during relearning are near zero. We tried several techniques to dampen them, which we describe below. Unfortunately they were outperformed by more straightforward techniques just relying on backpropagation. This failure is actually consistent with findings by \citet{finn_model-agnostic_2017}, who have shown that meta-learning which uses second-order derivatives does not perform any better than the simpler and cheaper first-order method (model-agnostic meta-learning).

\paragraph{Stream deactivation} Update of a weight is proportional to upstream activation and downstream gradient. So if we ensure that upstream activation is zero, then the update will be zero. Concretely, if we managed to deactivate the residual stream, we would prevent updates of first MLP layers which listen to residual stream activations. This could also be thought of as ensuring that \emph{nothing is represented} by the model--the activation is silent--so downstream modules are clueless about what is the context.

\paragraph{Misaligning second MLP layers from incoming gradients} We can look at the dual approach to silencing activations--silencing the backpropagating gradients. For a neuron in MLP, if its outgoing weights (one column in the second layer of the MLP) are orthogonal to the gradients flowing into the MLP, then this neuron's activation does not affect loss. This means that the gradient for this neuron's activation is zero, so we have stopped backpropagation flowing through this neuron.

\paragraph{Tweaking first MLP layers to dampen backpropagation} We know that the upstream gradient contributed by a given weight is its downstream gradient times the weight itself. Once we also know what is the sum of gradients contributed by all weights in a given module, we can then strategically tweak the weights to decrease magnitude of this summed gradient. Here in particular, we tried to tweak the first layers of the MLPs. This method has 3 variants: we can either aim to dampen gradient immediately upstream of this module, or the gradient before the layer norm, or even the gradient on the residual stream after the MLPs gradients are added into it.

\subsection{Direct weight edits}

Since we know that backpropagation tends to disable unwanted capabilities rather than removing them, we can try to more directly locate the weights responsible for these capabilities and ablate them. Unfortunately, here again we find that using backpropagation is more powerful in precisely locating where interventions are needed.
 
\paragraph{Ablating neurons based on activations} We can locate which MLP neurons are most active on the forget task, and at the same time least active on the retain task, and then just ablate them. This is similar to existing model editing methods \citep{wu_depn_2023,suau_whispering_2024}.

\paragraph{Ablating weights based on importance} We can also go more granular and look for ways to ablate weights. Here, we define weight relevance as: pre-weight activation times post-weight activation. Turns out this is \emph{dramatically better} than just ablating neurons, proving that intervening on neurons is not granular enough.

\paragraph{Fading backpropagation} We propose a technique that does rely on backpropagation, but at least tries to bias it to be more local--to prioritize effects nearer in the computation graph, rather than far downstream. To do so, we scale down the gradients added into the residual stream from each MLP (and optionally also from attention). The gradients passed through residual connections remain unscaled. Sadly, we have found no evidence of this working better than normal backpropagation.

\subsection{Erasing capabilities rather than drowning them out}

There is a possibility that our unlearning methods sometimes mask the unwanted capabilities by amplifying some other behavior, rather than actually erasing what we care about. So we can try to bias the methods to prefer erasing things rather than adding or amplifying.

\paragraph{Only shrinking weights} One thing to try is to only allow unlearning updates that shrink the magnitudes of model weights. So if an update has the same sign as the weight, it is zeroed out. It is not clear though that smaller weight magnitudes relate to erasing capabilities--for example a freshly initialized network contains many non-zero weights but no capabilities--only noise. So this heuristic of relying on weight magnitudes could definitely be refined.

\paragraph{Only shrinking activations} Another similar idea is to only allow weight updates which result in smaller \emph{activation} magnitudes. This is similar to "stream deactivation" discussed above, but here we do not rely on backpropagation and only care about activations immediately after a given weight (although this can be tweaked). We also do not actively aim to decrease these activations, just mask any updates that would grow them.

\paragraph{Selective logit loss} Finally, we can also limit which logits we try to shift. Normally, when we use cross-entropy loss applied after the final softmax, we can increase this loss either by decreasing logit of the correct token, or by \emph{increasing the logit for a token that is already highly active}. For this reason, unlearning methods will sometimes try to grow these other tokens which dominate the softmax, rather than just focusing on decreasing the logit of our unwanted token. 

To amend this, we can ignore the softmax layer, and focus just on bringing the unwanted token down. We have found that it helps to normalize by subtracting the average of all logits, to prevent incentivizing to decrease all logits. This is valid because output probabilities are invariant to shifting all logits by the same value. So the full equation for this more selective logit loss is: $$logit\_loss = logits_{correct\_id} - \sum_{i}^{} logits_i$$

We have found that it performs quite well, sometimes even outperforming cross-entropy loss or entropy loss. It is not very reliable across datasets though. In future work it is worth exploring deeper how to refine this technique.

\subsection{Making unlearning more selective}

We have seen great success in trying to make unlearning more selective. The culmination of this line of research was the Disruption Masking. Prior to it, we have tried many other ways of deciding which weights should be masked (e.g. quantiles and aggregating absolute values of gradients, which we describe below). We also tried multiple extensions (multiple LoRA adversaries, using weight consensus), but they provided no benefits.

\paragraph{Representation engineering retain loss} To make calculation of disruption more accurate, we can also augment the normal retain loss with a loss aiming to leave the model activations unchanged \cite{tamirisa_tamper-resistant_2024, zou_improving_2024}. In our setup we saw no improvement when using this activation loss, but maybe some effect will become visible after we also look at the performance on capability benchmarks.

\paragraph{Aggregating absolute values of gradients} Just adding the retaining gradients together, means that sometimes positive and negative ones will cancel. This means that if some weight increase disrupts performance in one context but helps in another, then it will be treated as neutral. If we want to be more conservative, we may disqualify such weights, based only on the fact that they disrupt in some contexts. To do so we can take the absolute value of retaining gradients and then aggregate them. Optionally we can also raise them to some power before aggregation.

It is possible to draw some parallels between Disruption Masking and A-GEM technique \citep{chaudhry_efficient_2019} from the continual learning field.
Similarly, the method in the continual learning field that corresponds to this aggregation of absolute values, would be Online-EWC \citep{schwarz_progress_2018}, which tries to estimate disruption on previous tasks by squaring the gradients.

By using absolute values of retaining gradients, we prevent canceling of helpful and unhelpful effects of changing a given weight, but now we also treat the helpful effects as something bad (since taking the absolute value inverts them). A compromise approach is to scale down the helpful effects so that they do not cancel the estimated disruption but also they are not counted as disruption themselves. We have found that this scaling works much better than just using absolute values, but it is still outperformed by Disruption Masking, which is also much more straightforward.

\paragraph{Disruption percentiles} Rather than attacking all weights where the signs of unlearning and retaining gradients agree, we could narrow down even more and attack just some small percent of weights least disruptive for retain performance and most disruptive for forget task performance. When we tune the percentile value which dictates how many weights we allow to attack, the optimal value oscillates around 50\%, meaning it is optimal to attack around half of all weights. This is more or less the same amount as when using Disruption Masking (where signs must agree, which happens around 50\% of the time). Given that just looking at gradient signs as the criterion is much simpler conceptually and requires no tuning, we got rid of using disruption percentiles.

\paragraph{Multiple adversaries} In our meta-unlearning approach, we mainly train just one adversary. This means the unlearning gradients we derive using it, may be idiosyncratic to that particular adversary. For this reason, we have also tried training multiple adversaries, which is actually the default in model-agnostic meta-learning. To save memory, each adversary was a LoRA adapter attached to the main model. When keeping the amount of compute constant in each variant, and tuning both the number of adversaries and the frequency of forking them, using multiple adversaries did not perform any better than just one (also a LoRA adapter). We speculate that splitting compute among multiple adversaries makes us do redundant work--each of the adversaries must travel through a similar learning trajectory.

\paragraph{Unlearning gradient accumulator} In the final MUDMAN method, we use an decaying accumulator to store retaining gradients. The same could be done for the unlearning gradients, so that during the masking step, we have a better idea of what is generally needed to break the forget set performance, not just break it on the current batch. (We also call this technique "forget momentum".) Interestingly, it turns out it is actually not needed. It also makes memory usage much higher, so we disabled it. This means that we need to accurately know which weights disrupt retain set performance (as we have shown in the main text), but knowing which ones break forget set performance does not need to be as accurate.

\paragraph{Weight consensus} When using multiple sources of unlearning gradients (for example when we have multiple adversaries or when we accumulate unlearning gradients over many batches), we can decide to attack only weights where there is a consensus that they are important for the forget task. This way we partially eliminate the role of luck and of adversary idiosyncrasies. To do so, we can instead of simply adding the gradients together, first raise them to some power smaller than one. This prioritizes consistency of update signs, and limits the influence of individual huge values. When we automatically tune this power, the optimal value tends to be around one, meaning that consensus is not important, and simply summing the gradients is sufficient. (We also allowed powers larger than one--similarly, they are worse.)

\subsection{Tweaking the meta-learning}

There are many ways to train the adversary models needed to perform model-agnostic meta-learning. We found the one described in Algorithm~\ref{alg:mudman} to be optimal. Here are some other ideas we tried:

\paragraph{LoRA adversaries} First of all, instead of using a full copy of the main model as our adversary, we could just attach a LoRA adapter and train this LoRA to do well on our forget task. This way we reuse main model weights, and the LoRA just serves as a small addition on top, used to reactivate dormant unwanted capabilities so that they can keep being unlearned. The performance of this variant was worse than using a full adversary, but sometimes it is competitive, so it may \emph{still be an option to consider when memory is a bottleneck}.

\paragraph{Adversary updates} One of the benefits of using LoRA adversaries, was that as the main model is updated, the adversary is naturally updated too (because it consists of a LoRA \emph{and} the main model weights underneath). This may mean that such adversary stays more up to date, or "in sync" with the main model, meaning we do not need to fork it as often. We could try to use this advantage also in the variant where not LoRA but a full adversary is used. Concretely, each time when we update the main model, we just apply the same update (optionally scaled down) to the adversary.

\paragraph{Adversary decay} Another way to keep the adversary more in sync with the main model, is to in each loop move its weights slightly closer to the main model. This may also be seen as a kind of regularization for the adversary. This was one of the most promising methods we found--for example on Figure~\ref{fig:failed_methods_and_cruelty} we can see that removing this mechanism dramatically harms unlearning performance on Python. However, we have found it to be unreliable across datasets, so we excluded it from final algorithm. We think refining this mechanism holds promise, though.

\paragraph{Locating unwanted circuits only once} To save compute we could even try removing meta-learning completely. But then, we get back to our initial problem--unwanted circuits are deactivated quickly (before they are fully erased), and so we cannot continue removing them with backpropagation. To remedy this, we tried to locate the unwanted circuit only once, using the initial model \emph{before any unlearning}. We go through the whole forget set (or some subset) and aggregate the gradients. The rest is the same as in MUDMAN--we apply this aggregated unlearning gradient, but only if its sign agrees with the retaining gradient (computed normally). This turns out to work really well, comparably to full meta-learning, but only at the beginning of the unlearning process. Later, it appears that this pre-computed unwanted circuit becomes too outdated.


\section{MUDMAN implementation in PyTorch}
\label{sec:mudman_implementation}

In Listing~\ref{lst:mudman_code} we show the core of the MUDMAN algorithm implemented in PyTorch.

In contrast to prior meta-unlearning algorithms, rather than training the adversary in an inner loop, we do everything in one loop and periodically fork the adversary. This simplification interleaves the main model and adversary updates more. We also focus on training only one adversary more deeply, rather than multiple ones shallowly. We found these changes to be beneficial, but we encourage future methods to explore these trade-offs more thoroughly.

In addition to inputs defined in Algorithm \ref{alg:mudman}, this code also needs \texttt{interven\_params} -- a list of parameters of the model to intervene on--in our case gate\_proj components of all the MLPs. Note that rather than defining a separate adversary model, we save memory by only storing adversarial weights for these \texttt{interven\_params}.

Concretely, rather than having \texttt{param.data}, we have \texttt{param.base\_data} and \texttt{param.adv\_data},
and for the inference we set \texttt{param.data} to point to one of these two. So in addition to the model, we need to store these additional weights, unlearning gradients and retain accumulators (also, only for \texttt{interven\_params}). This results in total memory usage of:
$$size(model) + 3 * size(interven\_params)$$
For intervening only on gate\_proj, this is less than regular training with SGD.

Another optimization is that we reuse the forward pass on forget batches, resulting in a total of just 5 forward or backward passes per loop. If cross-entropy is used as the unlearning loss, then reusing the backward pass is also possible.

\begin{figure*}
\centering
\lstdefinestyle{pythonstyle}{
    language=Python,
    backgroundcolor=\color{white},
    basicstyle=\footnotesize\ttfamily,
    breaklines=true,
    frame=single,
    numbers=left,
    numberstyle=\tiny\color{gray},
    keywordstyle=\color{blue},
    commentstyle=\color{orange},
    stringstyle=\color{red}
}
\begin{lstlisting}[style=pythonstyle, caption={Core of the MUDMAN Algorithm Implemented in PyTorch.}, label={lst:mudman_code}]
# Initialize retain grad accumulators
for p in interven_params:
    p.retain_acc = pt.zeros_like(p.data)
    p.base_data = p.data.clone().detach()

# Unlearning and retaining loop
for loop_num, (retain_batch, forget_batch) in enumerate(batch_pairs):
    if loop_num % fork_every_n_loops == 0:
        # Fork adversary
        for p in interven_params:
            p.adv_data = p.base_data.clone().detach()

    # Retain pass
    model.zero_grad()
    # Switch to base model
    for p in interven_params:
        p.data = p.base_data
    output = model(retain_batch)
    loss = cross_entropy_loss(output, retain_batch)
    loss.backward()
    for p in interven_params:
        # Update disruption scores
        p.retain_acc *= retain_momentum
        p.retain_acc += p.grad * (1 - retain_momentum)
        # Retain update
        p.base_data -= retaining_rate * p.retain_acc

    # Relearn the adversary
    model.zero_grad()
    # Switch to adversary
    for p in interven_params:
        p.data = p.adv_data
    output = model(forget_batch)
    loss = cross_entropy_loss(output, forget_batch)
    loss.backward(retain_graph=True)
    for p in interven_params:
        # Apply adversary update
        p.adv_data -= adv_lr * p.grad

    # Unlearning step with masking
    model.zero_grad()
    loss = unlearning_loss_fn(output, forget_batch)  # Reuse the output
    loss.backward()
    grad_norm = sum(p.grad.norm() ** 2 for p in interven_params) ** 0.5
    for p in interven_params:
        # Mask
        p.grad *= p.retain_acc.sign() == p.grad.sign()
        # Normalize & update
        p.base_data -= unlearning_rate / grad_norm * p.grad
\end{lstlisting}
\end{figure*}

\clearpage
\clearpage
\section{Hyperparameter searches}
\label{sec:hyperparameter_searches}

Each bar in Figures~\ref{fig:ablation_study}~and~\ref{fig:wmdp_accuracy} corresponds to one Optuna hyperparameter search. In Table~\ref{tab:search_configs} we report for each model: the number of trials, unlearning steps, relearning steps, and approximate total time of one search on one Nvidia L40 GPU. In "WMDP" row we also report these values for the final experiment--unlearning and relearning on Pile-Bio, and trying to minimize WMDP-Bio accuracy.

By unlearning and relearning steps, we do not mean the number of algorithm loops, but the total number of forward and backward passes. This way, we ensure that each run used roughly the same amount of compute, regardless of the method used.

\begin{table}[h]
\caption{Hyperparameter search configurations for each model. \emph{Trials} indicates the number of Optuna trials, \emph{Unl.} and \emph{Rel.} show the number of steps in each phase, and \emph{Time} is the approximate total duration of the search. (Main paper experiments use 5 searches per one method comparison, so multiply the "Time" value times 5 for the total duration of a method comparison.)}
\centering
\begin{tabular}{lrrrr}
\hline
\textbf{Model} & \textbf{Trials} & \textbf{Unl.} & \textbf{Rel.} & \textbf{Time} \\
\hline
Llama & 500 & 120 & 120 & 7h \\
Smol & 500 & 300 & 300 & 9h \\
pythia & 800 & 300 & 300 & 4h \\
\hline
WMDP & 300 & 480 & 240 & 20h \\
\hline
\end{tabular}
\label{tab:search_configs}
\end{table}

We always use the same relearning process: SGD with a learning rate of 1e-3, and in the case of WMDP, 3e-4. We also tried using LoRAs for relearning, but that resulted in unpredictable results--probably some LoRAs just have a lucky initialization. This makes the comparison of methods too noisy, so we fixed on only using SGD relearning.

We made sure that hyperparameter search ranges are wide enough to cover the best values. For exact ranges used in each search, you can look at \href{https://github.com/filyp/MUDMAN/tree/main/configs}{configuration files} named \texttt{ablations\_and\_loss*} and \texttt{wmdp7}.

\subsection{Maximizing forget loss searches}

Note that to produce Figure~\ref{fig:ablation_study}, we use a MUDMAN version where the adversary's weights are moved slightly towards the main model weights in each step and the strength of this is tuned (adversary decay). This is something which we later abandoned (Algorithm~\ref{alg:mudman} does not have it) after realising it does not help and only complicates the algorithm.

We tune unlearning\_rate ($\alpha_{unlearning}$ in Algorithm~\ref{alg:mudman}), retaining\_rate ($\alpha_{retaining}$), adv\_lr ($\alpha_{adv}$), retain\_momentum ($\mu$), adv\_decay (later removed from Algorithm~\ref{alg:mudman}), and fork\_every\_n\_loops (how often we fork the adversary).

\subsection{Minimizing WMDP accuracy searches}

We only attack Llama-3.2-1B, because other models have accuracy not better than random guessing (25\%), while Llama-3.2-1B has about 45\%.



To achieve significant decrease in WMDP accuracy, we each trial is 3x longer than before. This would make searches too long, so we only tune the unlearning\_rate and retaining\_rate, which are the two most important hyperparameters.

Previous automatic searches inform the choice of other hyperparameters. When the retain loss exceeds a predefined threshold (initial retain loss + 0.05), we pause the unlearning updates while still doing the retaining updates, until retain loss goes back below the threshold. If retain loss exceeds a higher threshold (initial retain loss + 0.1) we terminate and reject the trial. This ensures each method has almost the same impact on the retain set performance.

\subsection{Detailed Optuna plots}

\begin{figure*}[ht]
    \centering
    \begin{tabular}{cc}
        \includegraphics[height=0.445\textheight]{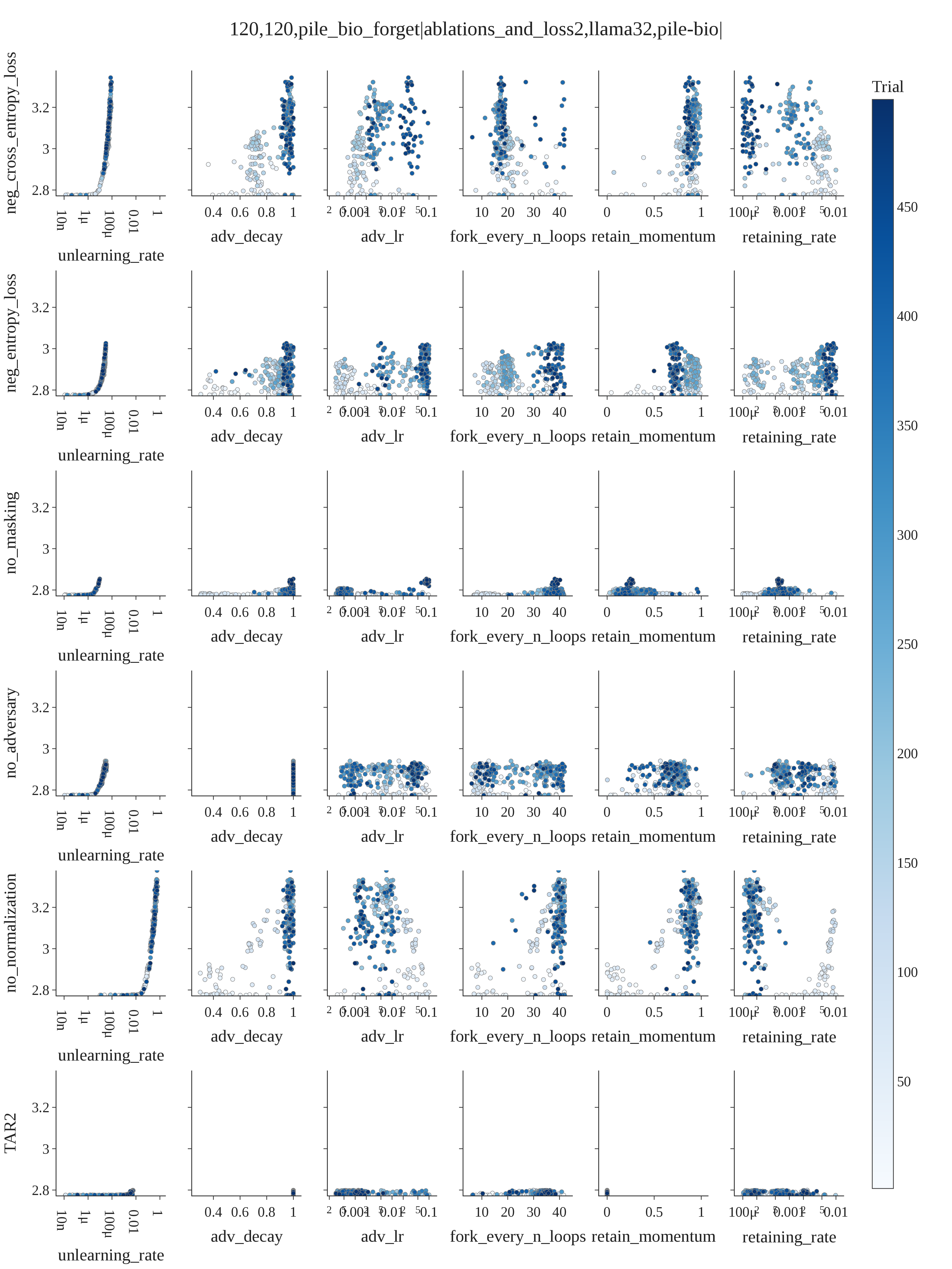} &
        \includegraphics[height=0.445\textheight]{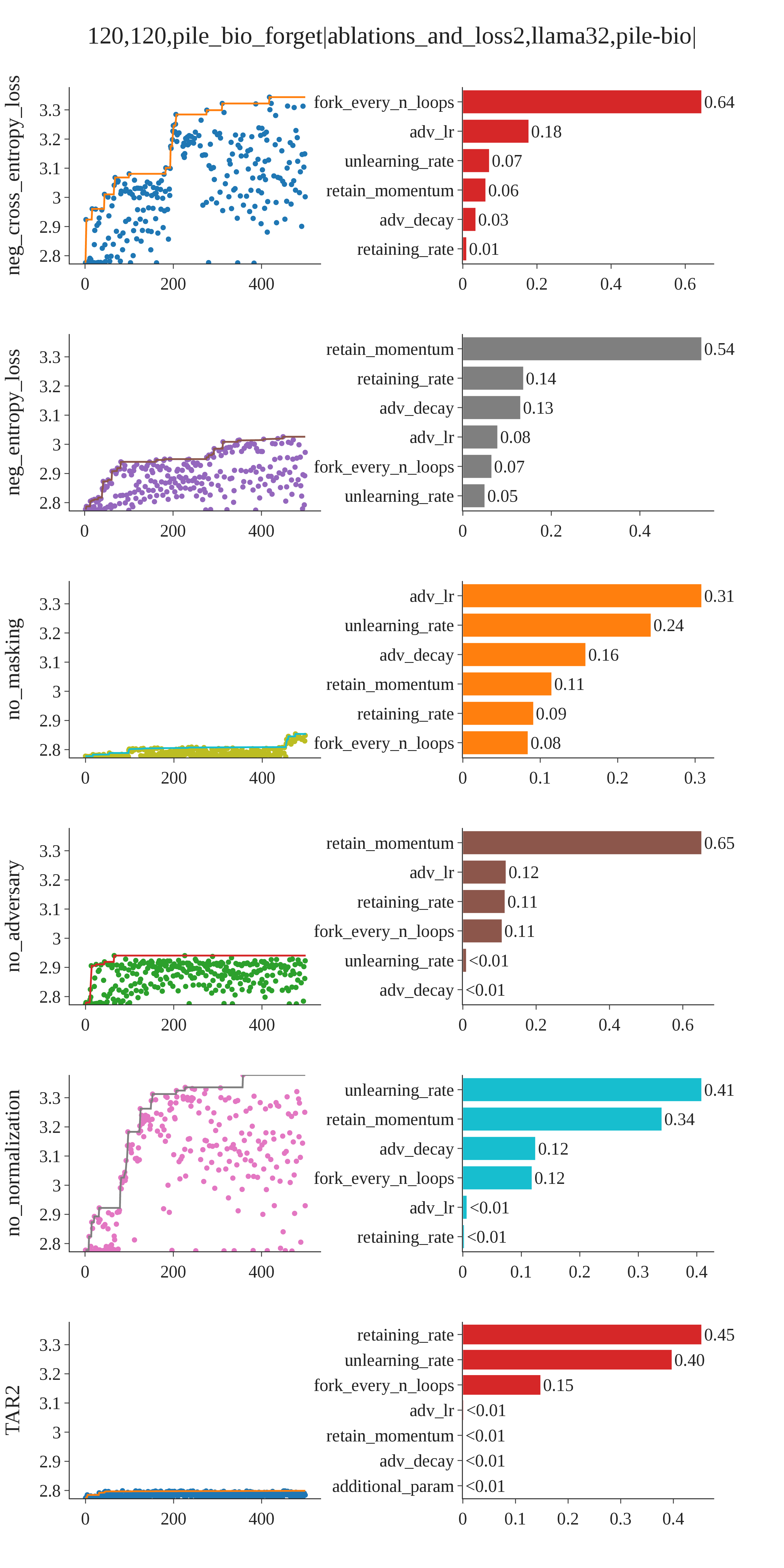} \\
        \includegraphics[height=0.445\textheight]{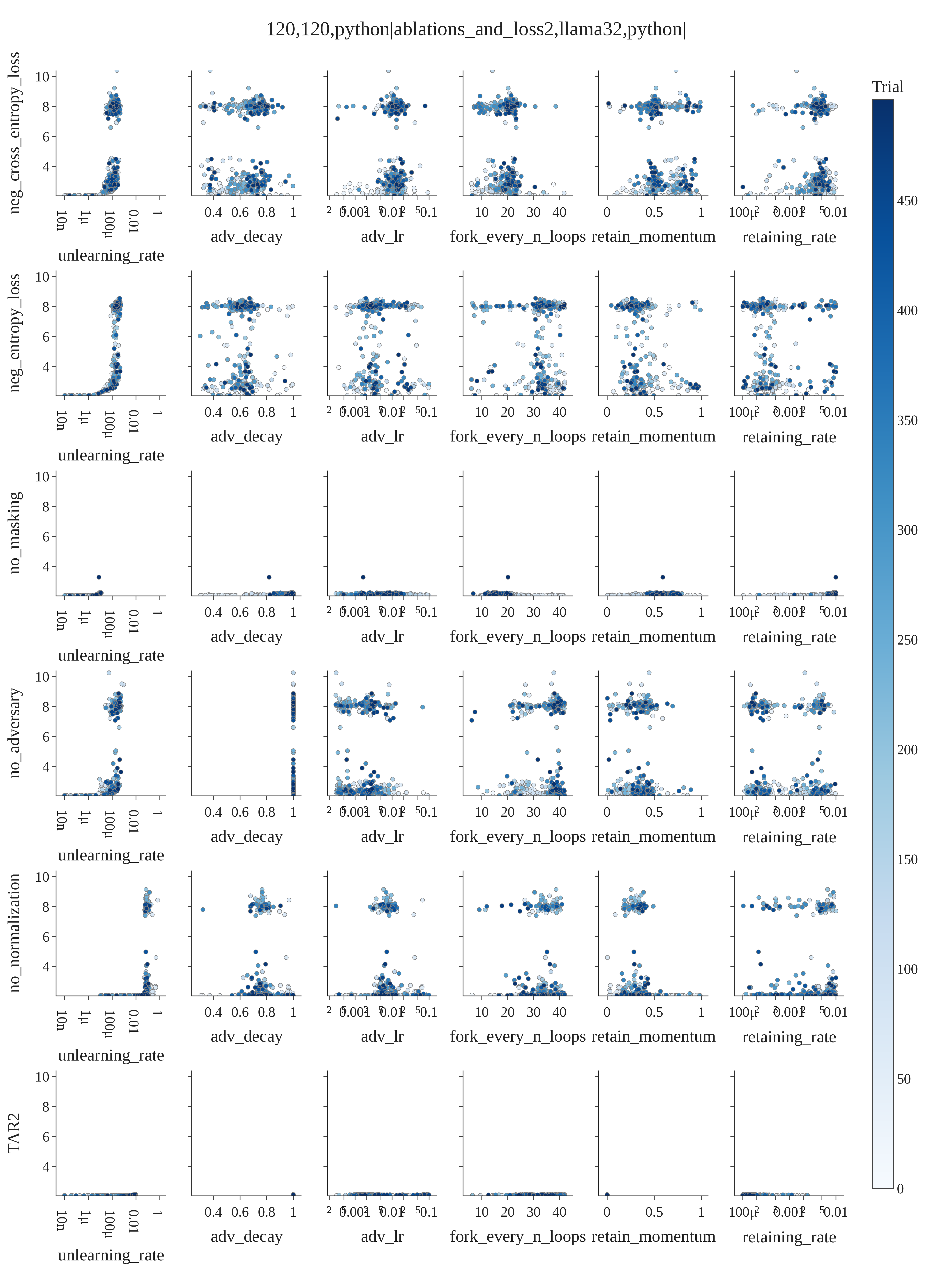} &
        \includegraphics[height=0.445\textheight]{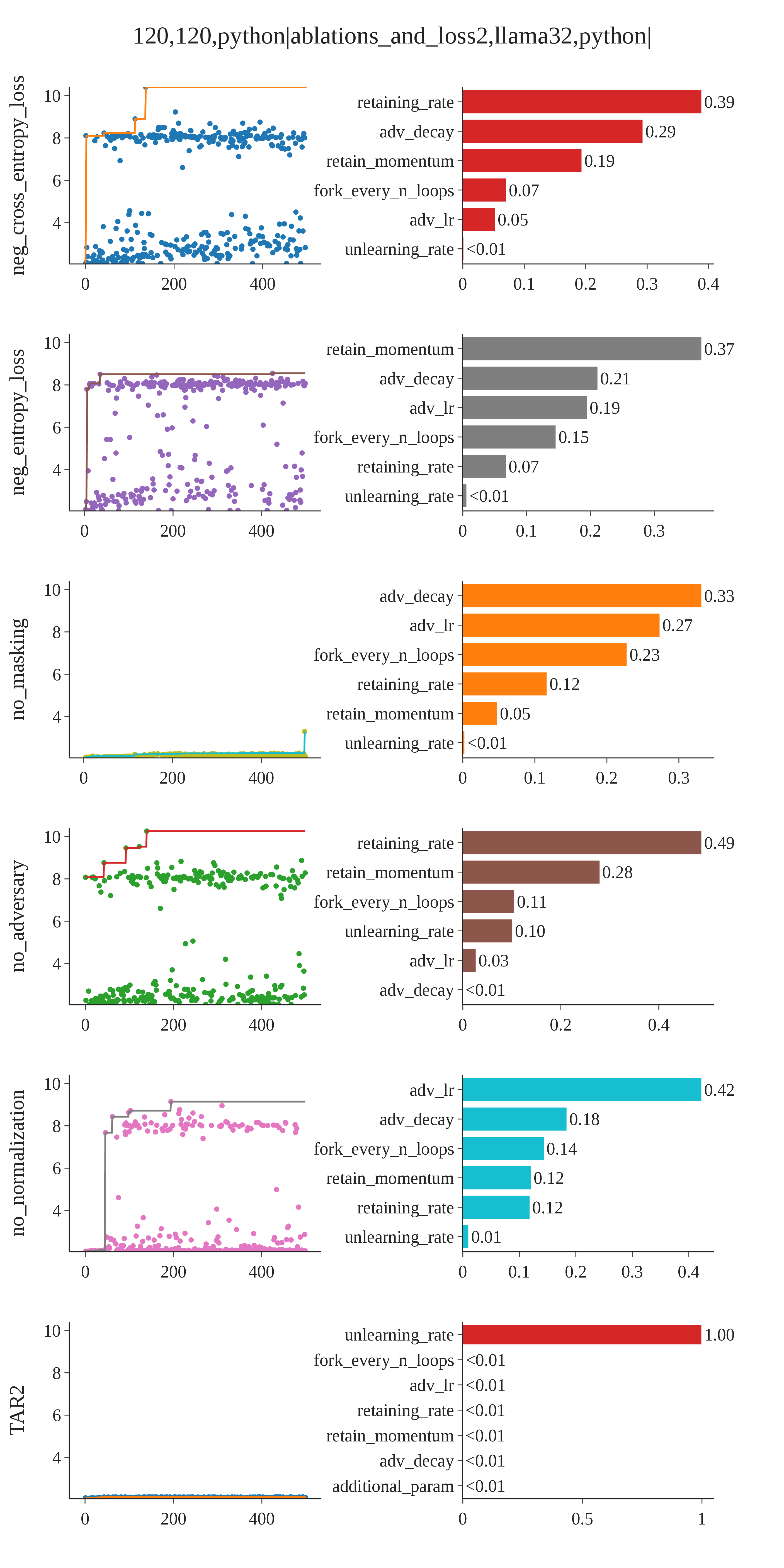}
    \end{tabular}
    \caption{Hyperparameter optimization results for Llama-3.2-1B. Top: Pile-Bio dataset. Bottom: Python dataset. Left: Forget loss depending on each hyperparameter. Right: Optimization history and hyperparameter importance.}
    \label{fig:optuna_llama}
\end{figure*}

\begin{figure*}[ht]
    \centering
    \begin{tabular}{cc}
        \includegraphics[height=0.445\textheight]{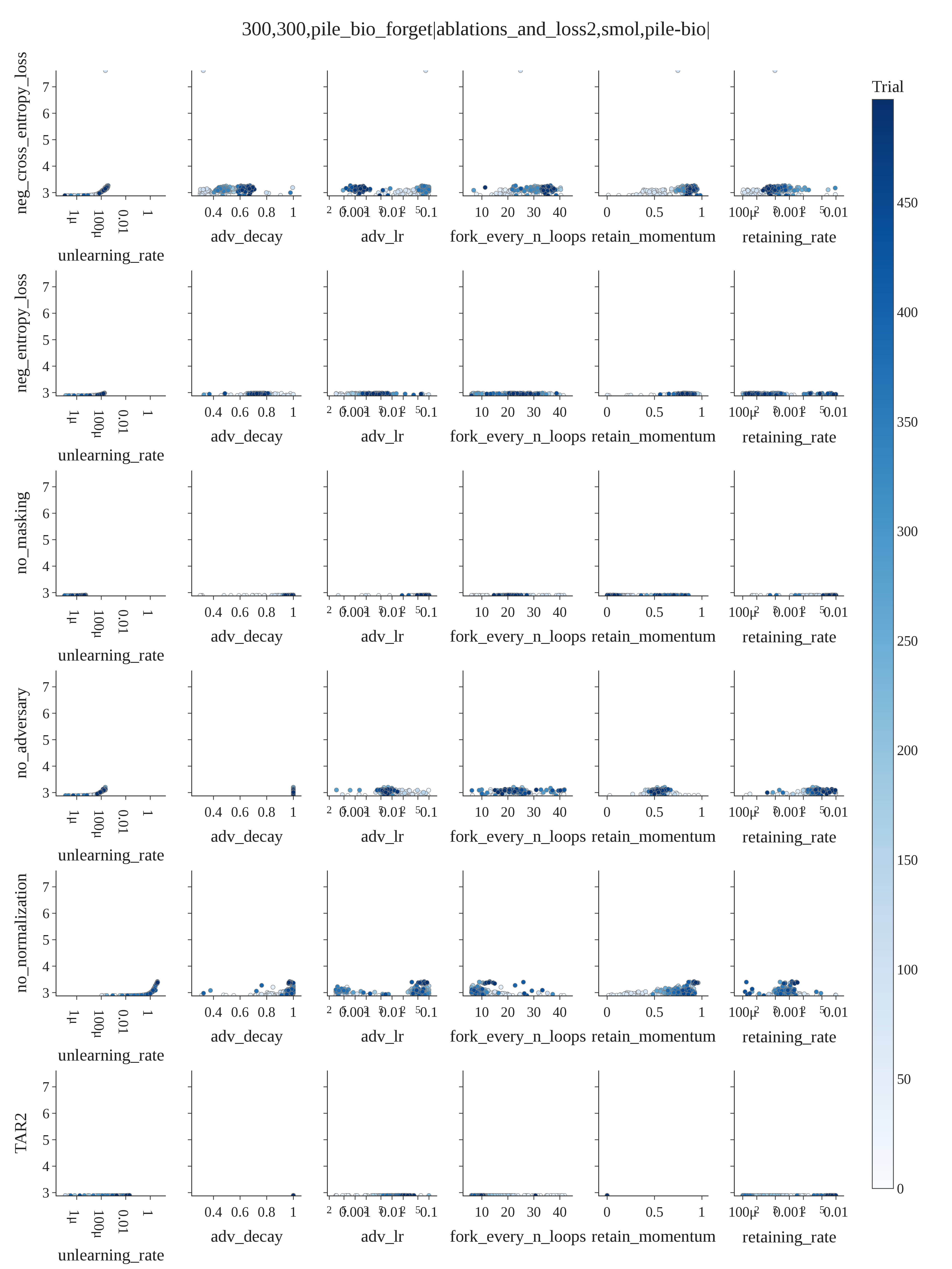} &
        \includegraphics[height=0.445\textheight]{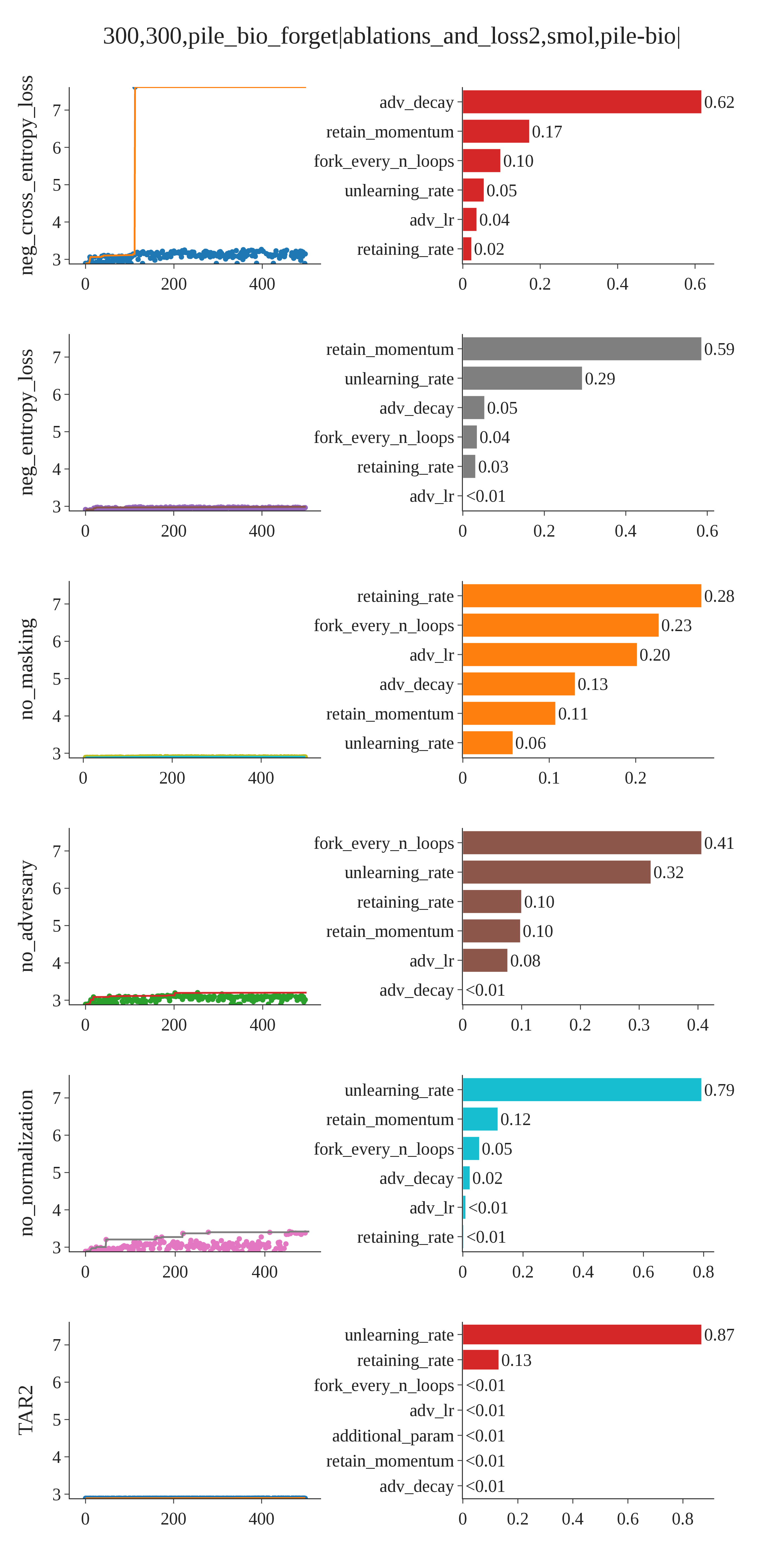} \\
        \includegraphics[height=0.445\textheight]{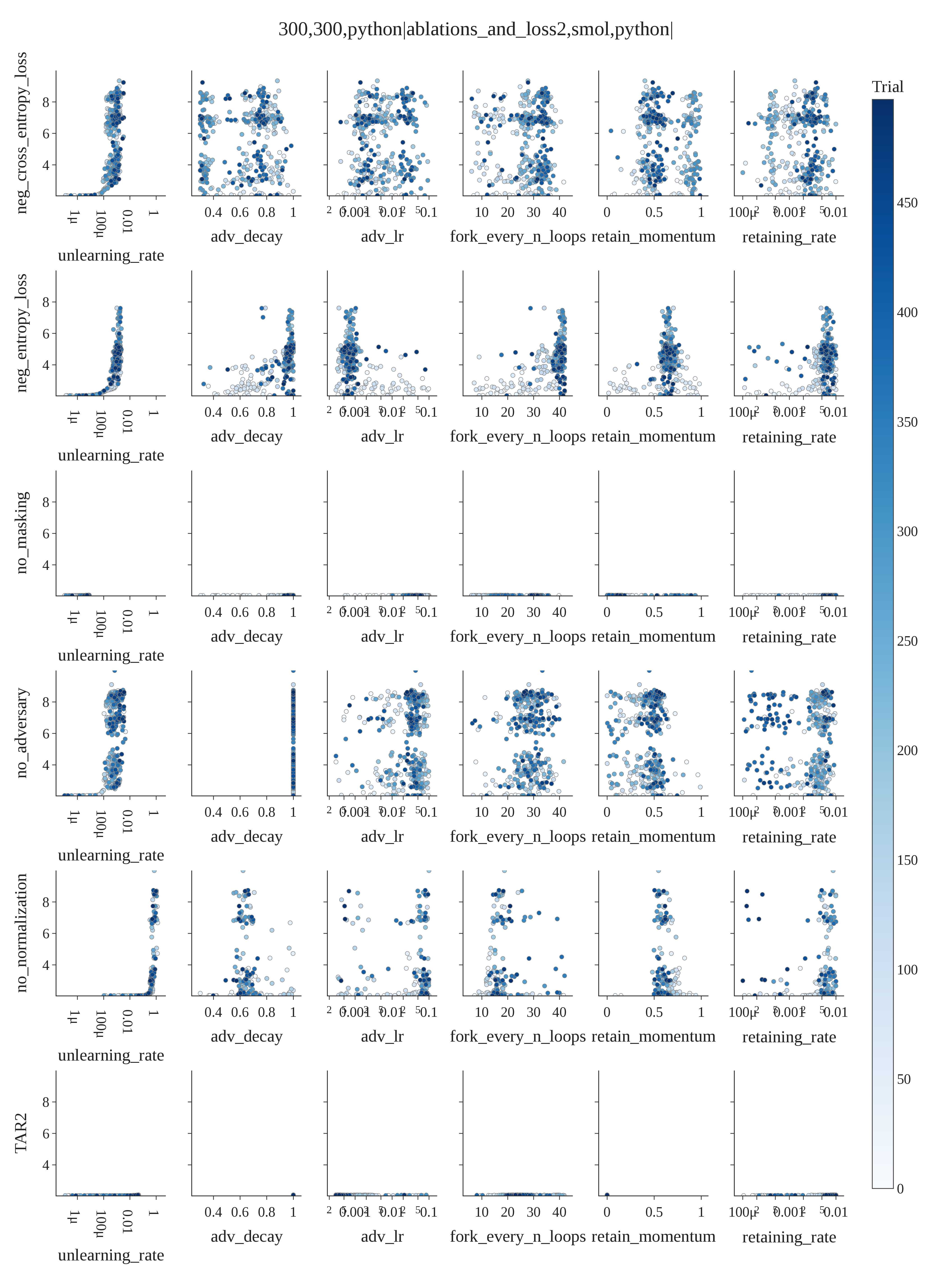} &
        \includegraphics[height=0.445\textheight]{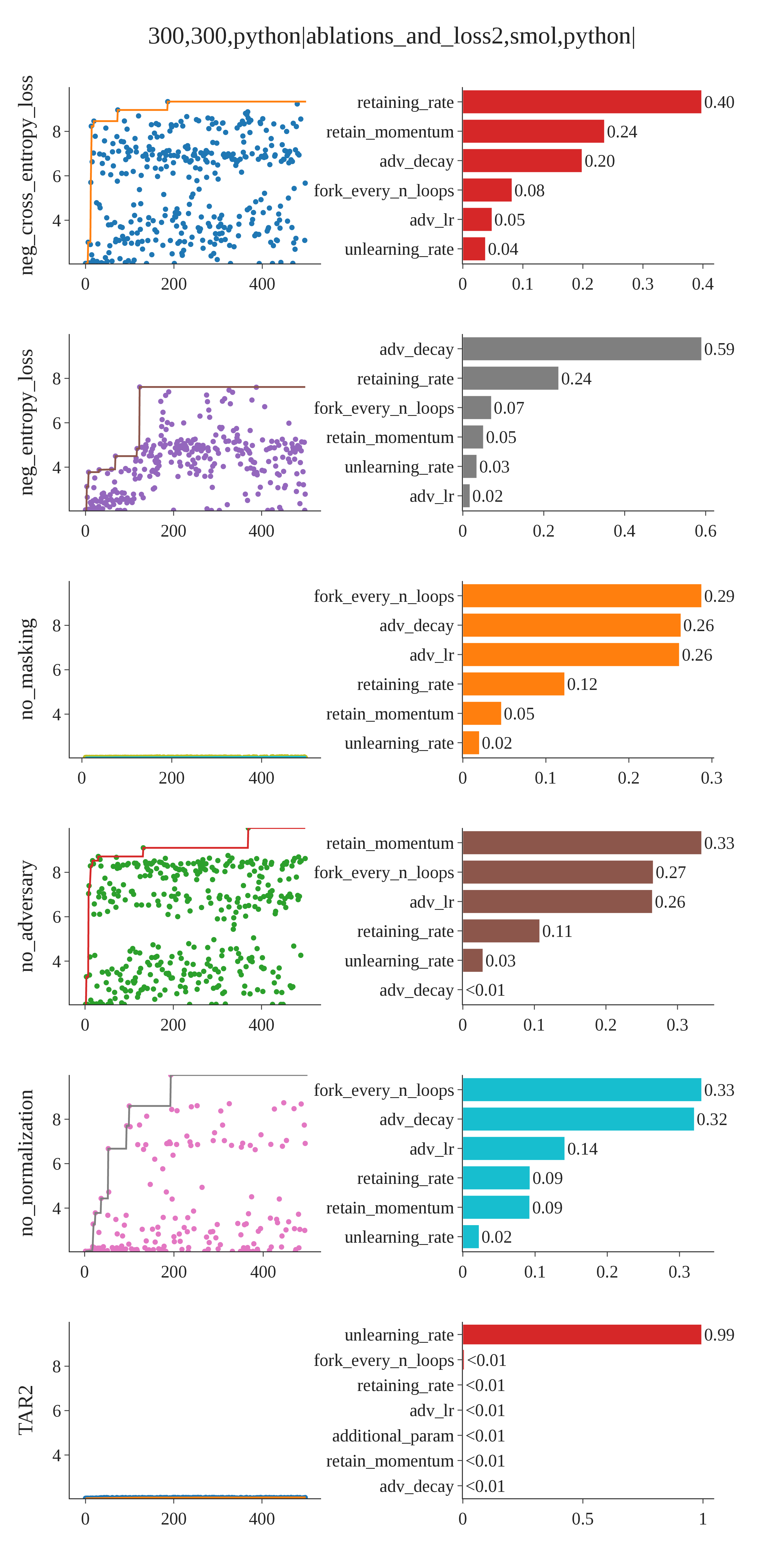}
    \end{tabular}
    \caption{Hyperparameter optimization results for SmolLM-135M. Top: Pile-Bio dataset. Bottom: Python dataset. Left: Forget loss depending on each hyperparameter. Right: Optimization history and hyperparameter importance.}
    \label{fig:optuna_smol}
\end{figure*}

\begin{figure*}[ht]
    \centering
    \begin{tabular}{cc}
        \includegraphics[height=0.445\textheight]{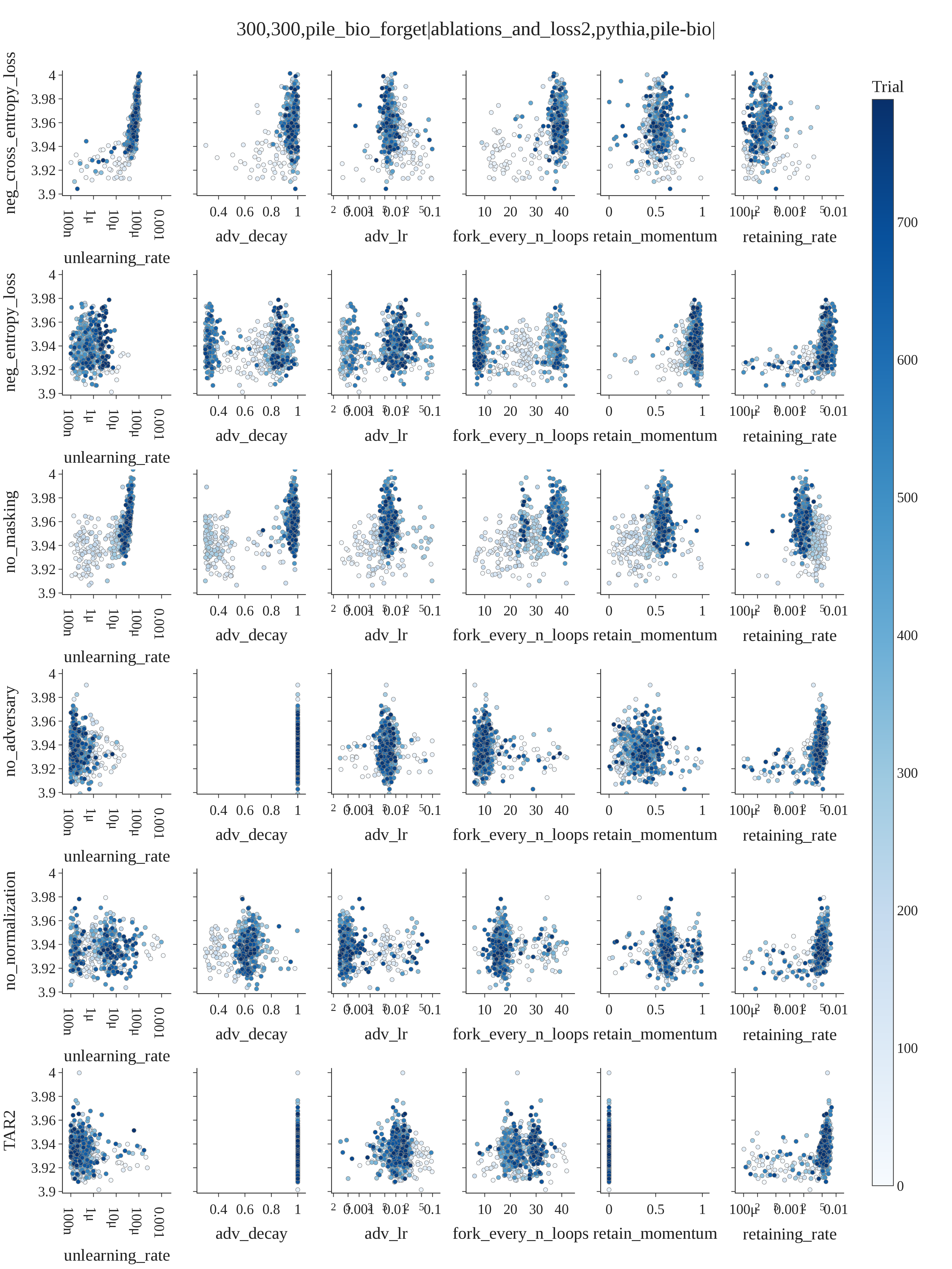} &
        \includegraphics[height=0.445\textheight]{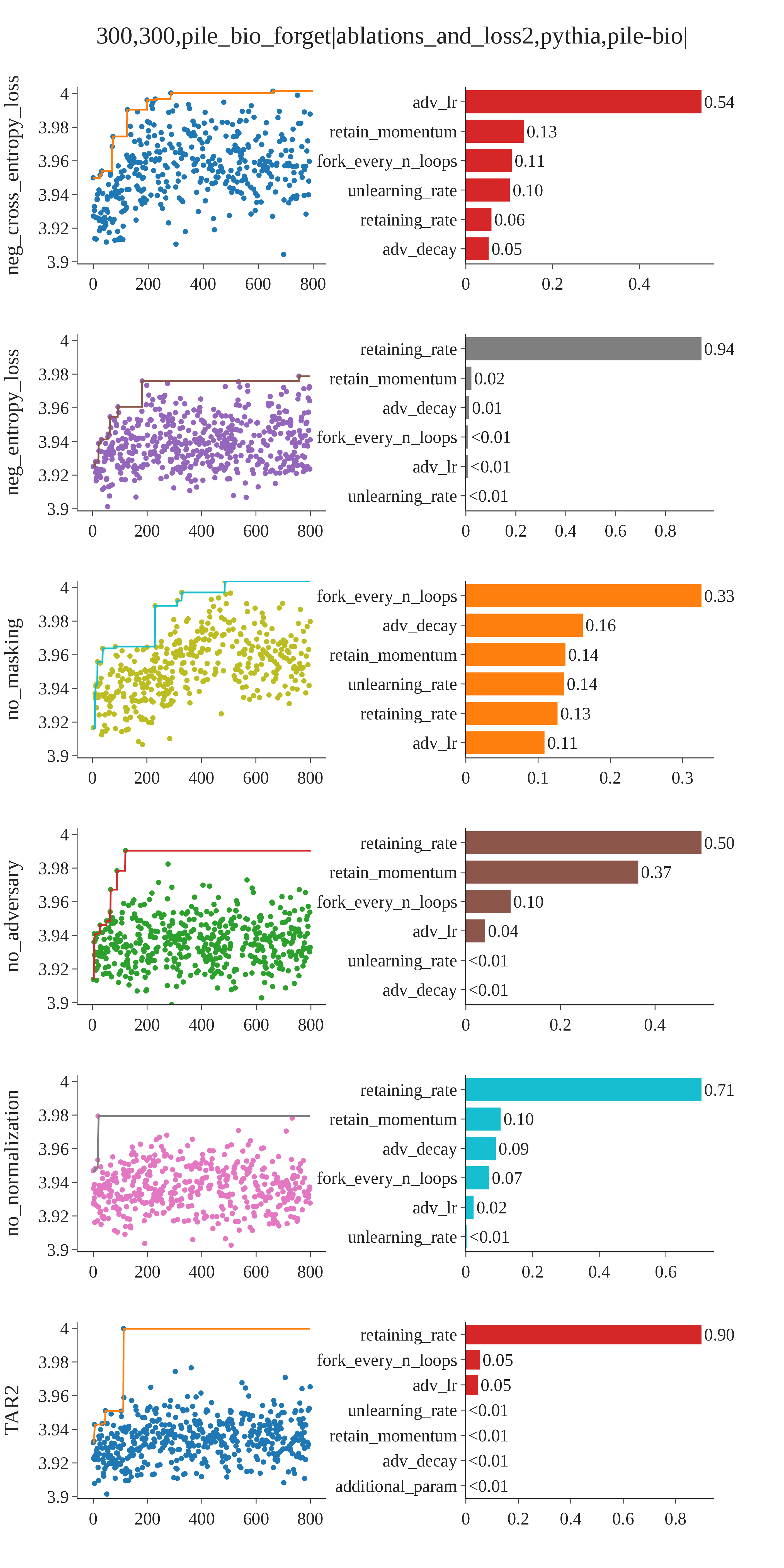} \\
        \includegraphics[height=0.445\textheight]{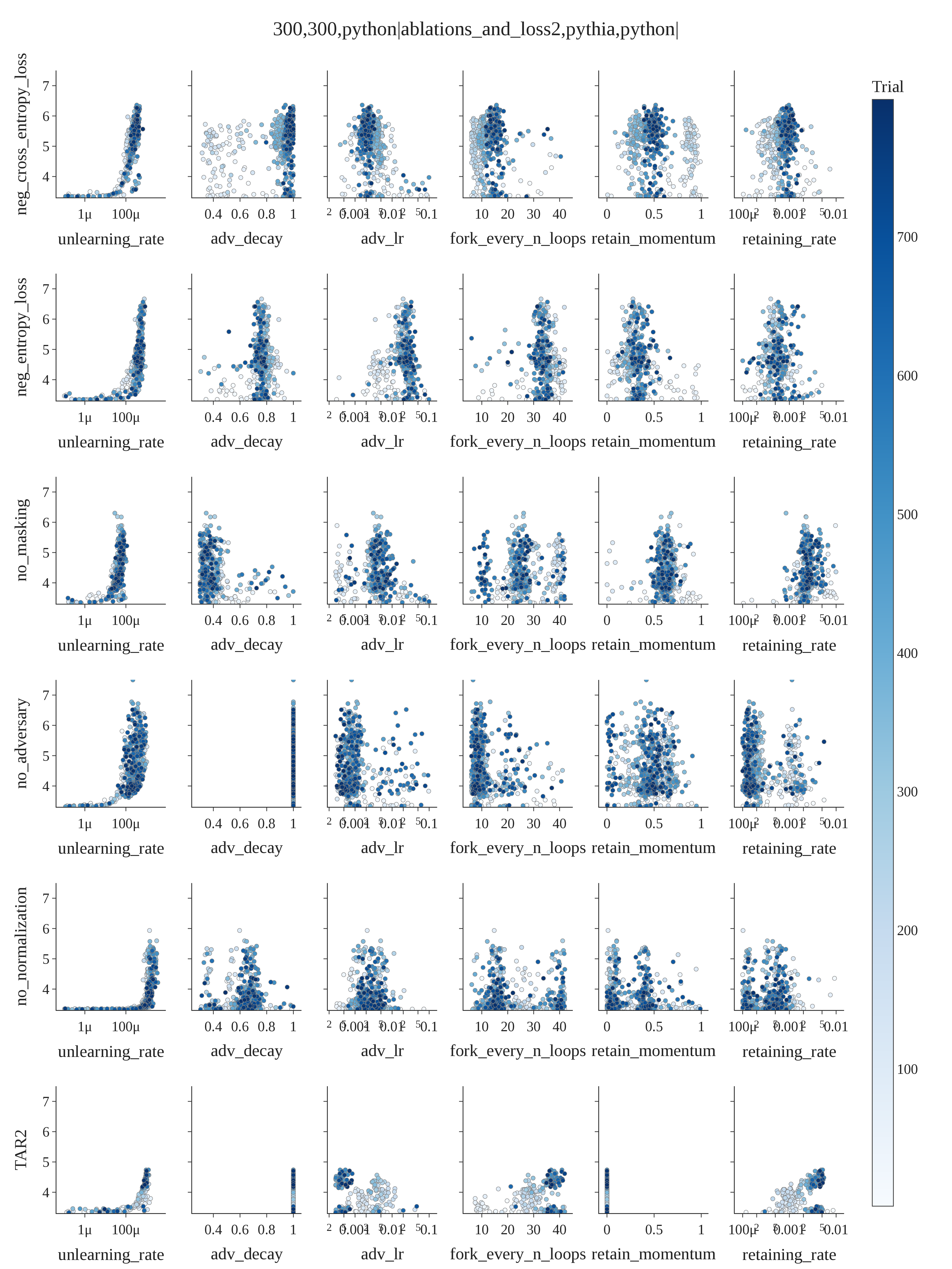} &
        \includegraphics[height=0.445\textheight]{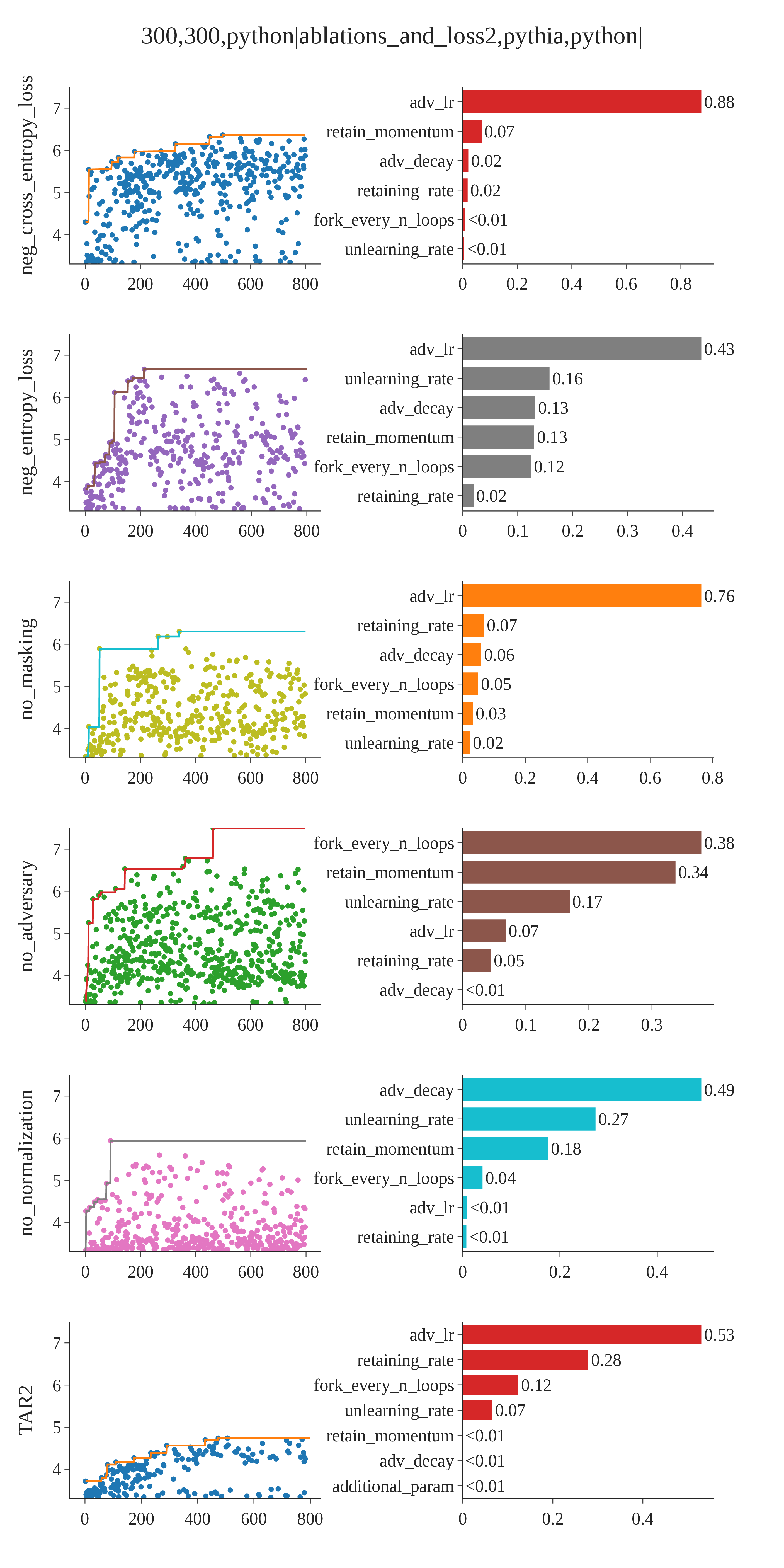}
    \end{tabular}
    \caption{Hyperparameter optimization results for pythia-14m. Top: Pile-Bio dataset. Bottom: Python dataset. Left: Forget loss depending on each hyperparameter. Right: Optimization history and hyperparameter importance.}
    \label{fig:optuna_pythia}
\end{figure*}

\begin{figure*}[ht]
    \centering
    \begin{tabular}{cc}
        \includegraphics[height=0.8\textheight]{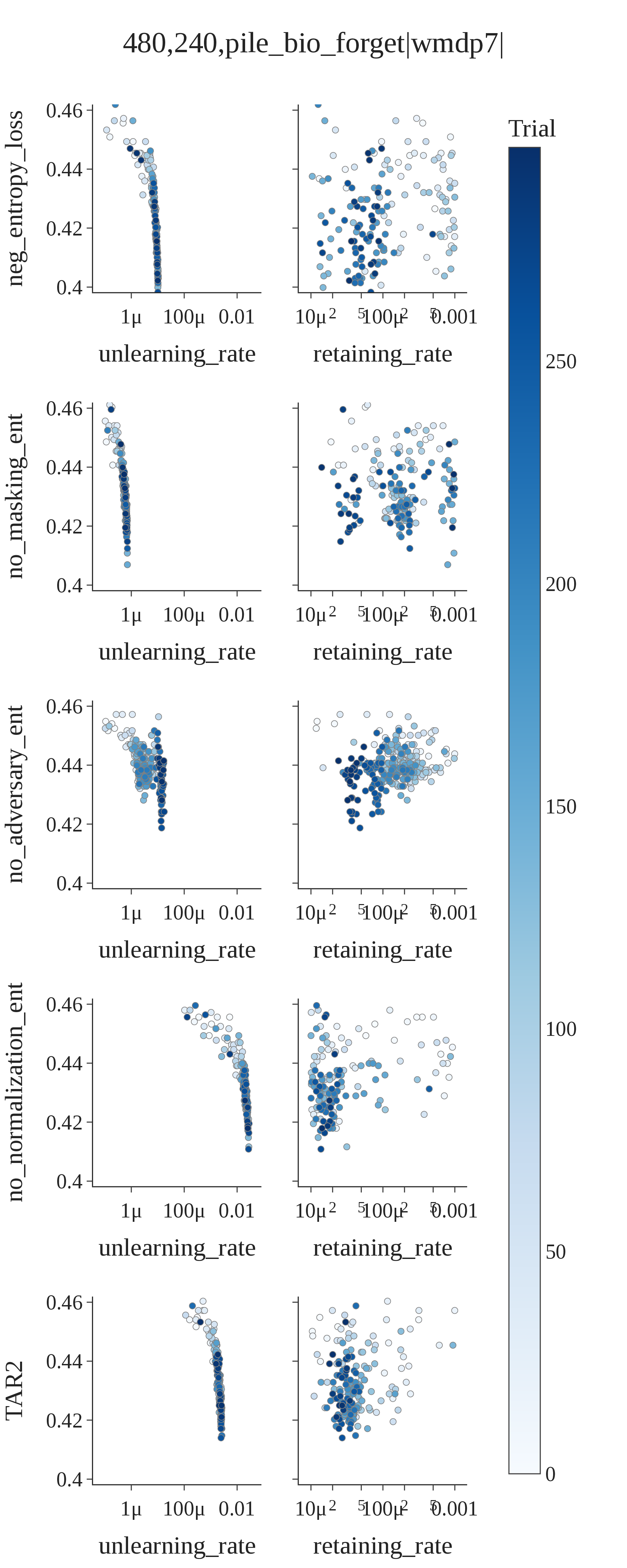} &
        \includegraphics[height=0.8\textheight]{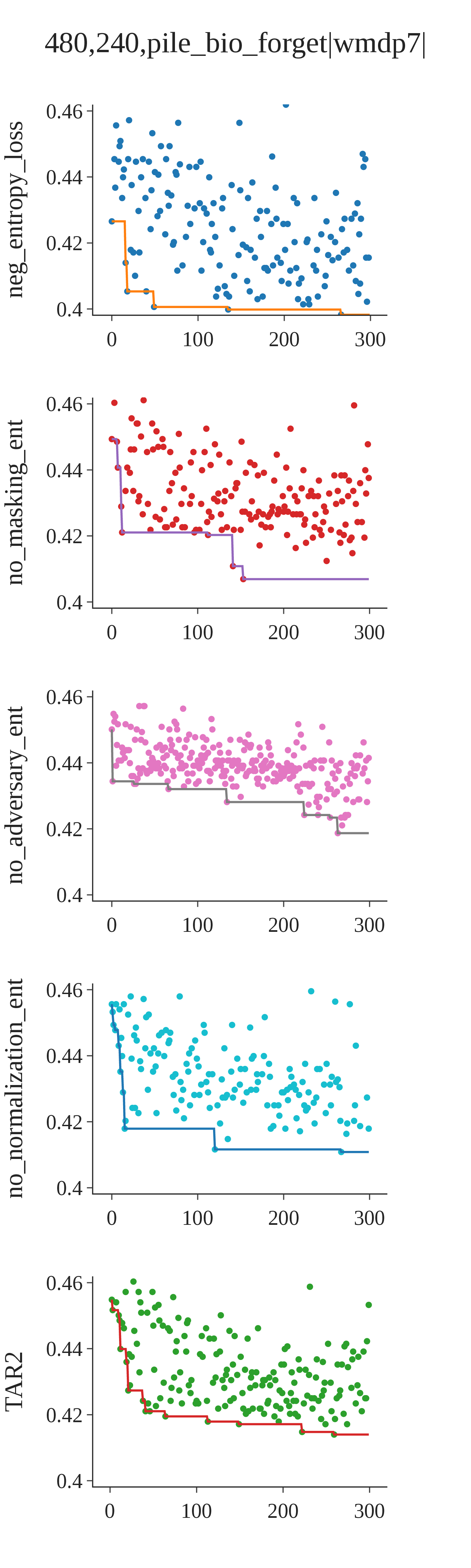}
    \end{tabular}
    \caption{Hyperparameter optimization results for WMDP accuracy minimization. Left: Forget loss depending on each hyperparameter. Right: Optimization history.}
    \label{fig:wmdp_optimization}
\end{figure*}

Finally, for each Optuna search, we provide detailed plots, containing per-trial results. In Figures~\ref{fig:optuna_llama}~--~\ref{fig:wmdp_optimization} each row corresponds to one search and each point to one trial. On the left you can see how performance (forget loss after relearning, or WMDP accuracy) depends on the values of each hyperparameter. The color marks the order of trials, with dark blue trials happening late in the search. On the right you can see the optimization history (how performance increased as the search progressed) and if more than one hyperparameter was used, you can also see the estimates of hyperparameter importance produced by Optuna. Some clouds of points do not span the full range, because pruned trials are not shown, and also no\_masking and no\_normalization searches can use different unlearning\_rate ranges.


\clearpage


\newpage
\section*{NeurIPS Paper Checklist}

\begin{enumerate}

\item {\bf Claims}
    \item[] Question: Do the main claims made in the abstract and introduction accurately reflect the paper's contributions and scope?
    \item[] Answer: \answerYes{} 
    \item[] Justification: The claimed improvement over SOTA is comes from the experiments desribed in the paper. All components claimed to be useful, are indeed proved to be necessary by the ablation study.
    \item[] Guidelines:
    \begin{itemize}
        \item The answer NA means that the abstract and introduction do not include the claims made in the paper.
        \item The abstract and/or introduction should clearly state the claims made, including the contributions made in the paper and important assumptions and limitations. A No or NA answer to this question will not be perceived well by the reviewers. 
        \item The claims made should match theoretical and experimental results, and reflect how much the results can be expected to generalize to other settings. 
        \item It is fine to include aspirational goals as motivation as long as it is clear that these goals are not attained by the paper. 
    \end{itemize}

\item {\bf Limitations}
    \item[] Question: Does the paper discuss the limitations of the work performed by the authors?
    \item[] Answer: \answerYes{} 
    \item[] Justification: We include a separate "Limitations" section in the paper. We discuss there all the aspects that we had to exclude to not overcomplicate the paper. Computational efficiency has been discussed in Appendix~\ref{sec:mudman_implementation}.
    \item[] Guidelines:
    \begin{itemize}
        \item The answer NA means that the paper has no limitation while the answer No means that the paper has limitations, but those are not discussed in the paper. 
        \item The authors are encouraged to create a separate "Limitations" section in their paper.
        \item The paper should point out any strong assumptions and how robust the results are to violations of these assumptions (e.g., independence assumptions, noiseless settings, model well-specification, asymptotic approximations only holding locally). The authors should reflect on how these assumptions might be violated in practice and what the implications would be.
        \item The authors should reflect on the scope of the claims made, e.g., if the approach was only tested on a few datasets or with a few runs. In general, empirical results often depend on implicit assumptions, which should be articulated.
        \item The authors should reflect on the factors that influence the performance of the approach. For example, a facial recognition algorithm may perform poorly when image resolution is low or images are taken in low lighting. Or a speech-to-text system might not be used reliably to provide closed captions for online lectures because it fails to handle technical jargon.
        \item The authors should discuss the computational efficiency of the proposed algorithms and how they scale with dataset size.
        \item If applicable, the authors should discuss possible limitations of their approach to address problems of privacy and fairness.
        \item While the authors might fear that complete honesty about limitations might be used by reviewers as grounds for rejection, a worse outcome might be that reviewers discover limitations that aren't acknowledged in the paper. The authors should use their best judgment and recognize that individual actions in favor of transparency play an important role in developing norms that preserve the integrity of the community. Reviewers will be specifically instructed to not penalize honesty concerning limitations.
    \end{itemize}

\item {\bf Theory assumptions and proofs}
    \item[] Question: For each theoretical result, does the paper provide the full set of assumptions and a complete (and correct) proof?
    \item[] Answer: \answerNA{} 
    \item[] Justification: There are no theorems that we introduce.
    \item[] Guidelines:
    \begin{itemize}
        \item The answer NA means that the paper does not include theoretical results. 
        \item All the theorems, formulas, and proofs in the paper should be numbered and cross-referenced.
        \item All assumptions should be clearly stated or referenced in the statement of any theorems.
        \item The proofs can either appear in the main paper or the supplemental material, but if they appear in the supplemental material, the authors are encouraged to provide a short proof sketch to provide intuition. 
        \item Inversely, any informal proof provided in the core of the paper should be complemented by formal proofs provided in appendix or supplemental material.
        \item Theorems and Lemmas that the proof relies upon should be properly referenced. 
    \end{itemize}

    \item {\bf Experimental result reproducibility}
    \item[] Question: Does the paper fully disclose all the information needed to reproduce the main experimental results of the paper to the extent that it affects the main claims and/or conclusions of the paper (regardless of whether the code and data are provided or not)?
    \item[] Answer: \answerYes{} 
    \item[] Justification: We provide the natural language intuitive description of the algorithm, and also the pseudocode, and also a minimal implementation in PyTorch in Appendix~\ref{sec:mudman_implementation}, and a link to the repository with the actual code. We also explain in detail the setup used to compare different methods in Section~\ref{sec:methodology} and in Appendix~\ref{sec:hyperparameter_searches}.
    \item[] Guidelines:
    \begin{itemize}
        \item The answer NA means that the paper does not include experiments.
        \item If the paper includes experiments, a No answer to this question will not be perceived well by the reviewers: Making the paper reproducible is important, regardless of whether the code and data are provided or not.
        \item If the contribution is a dataset and/or model, the authors should describe the steps taken to make their results reproducible or verifiable. 
        \item Depending on the contribution, reproducibility can be accomplished in various ways. For example, if the contribution is a novel architecture, describing the architecture fully might suffice, or if the contribution is a specific model and empirical evaluation, it may be necessary to either make it possible for others to replicate the model with the same dataset, or provide access to the model. In general. releasing code and data is often one good way to accomplish this, but reproducibility can also be provided via detailed instructions for how to replicate the results, access to a hosted model (e.g., in the case of a large language model), releasing of a model checkpoint, or other means that are appropriate to the research performed.
        \item While NeurIPS does not require releasing code, the conference does require all submissions to provide some reasonable avenue for reproducibility, which may depend on the nature of the contribution. For example
        \begin{enumerate}
            \item If the contribution is primarily a new algorithm, the paper should make it clear how to reproduce that algorithm.
            \item If the contribution is primarily a new model architecture, the paper should describe the architecture clearly and fully.
            \item If the contribution is a new model (e.g., a large language model), then there should either be a way to access this model for reproducing the results or a way to reproduce the model (e.g., with an open-source dataset or instructions for how to construct the dataset).
            \item We recognize that reproducibility may be tricky in some cases, in which case authors are welcome to describe the particular way they provide for reproducibility. In the case of closed-source models, it may be that access to the model is limited in some way (e.g., to registered users), but it should be possible for other researchers to have some path to reproducing or verifying the results.
        \end{enumerate}
    \end{itemize}

\item {\bf Open access to data and code}
    \item[] Question: Does the paper provide open access to the data and code, with sufficient instructions to faithfully reproduce the main experimental results, as described in supplemental material?
    \item[] Answer: \answerYes{} 
    \item[] Justification: We link to the repository and include instructions how to run the experiments. We also describe the datasets used.
    \item[] Guidelines:
    \begin{itemize}
        \item The answer NA means that paper does not include experiments requiring code.
        \item Please see the NeurIPS code and data submission guidelines (\url{https://nips.cc/public/guides/CodeSubmissionPolicy}) for more details.
        \item While we encourage the release of code and data, we understand that this might not be possible, so “No” is an acceptable answer. Papers cannot be rejected simply for not including code, unless this is central to the contribution (e.g., for a new open-source benchmark).
        \item The instructions should contain the exact command and environment needed to run to reproduce the results. See the NeurIPS code and data submission guidelines (\url{https://nips.cc/public/guides/CodeSubmissionPolicy}) for more details.
        \item The authors should provide instructions on data access and preparation, including how to access the raw data, preprocessed data, intermediate data, and generated data, etc.
        \item The authors should provide scripts to reproduce all experimental results for the new proposed method and baselines. If only a subset of experiments are reproducible, they should state which ones are omitted from the script and why.
        \item At submission time, to preserve anonymity, the authors should release anonymized versions (if applicable).
        \item Providing as much information as possible in supplemental material (appended to the paper) is recommended, but including URLs to data and code is permitted.
    \end{itemize}

\item {\bf Experimental setting/details}
    \item[] Question: Does the paper specify all the training and test details (e.g., data splits, hyperparameters, how they were chosen, type of optimizer, etc.) necessary to understand the results?
    \item[] Answer: \answerYes{} 
    \item[] Justification: We discuss them in Section~\ref{sec:methodology} and in Appendix~\ref{sec:hyperparameter_searches}. We describe the hyperparameter searches, and link to the configuration files.
    \item[] Guidelines:
    \begin{itemize}
        \item The answer NA means that the paper does not include experiments.
        \item The experimental setting should be presented in the core of the paper to a level of detail that is necessary to appreciate the results and make sense of them.
        \item The full details can be provided either with the code, in appendix, or as supplemental material.
    \end{itemize}

\item {\bf Experiment statistical significance}
    \item[] Question: Does the paper report error bars suitably and correctly defined or other appropriate information about the statistical significance of the experiments?
    \item[] Answer: \answerYes{} 
    \item[] Justification: We have a large number of runs for each hyperparameter search, so we report standard errors calculated over these runs.
    \item[] Guidelines:
    \begin{itemize}
        \item The answer NA means that the paper does not include experiments.
        \item The authors should answer "Yes" if the results are accompanied by error bars, confidence intervals, or statistical significance tests, at least for the experiments that support the main claims of the paper.
        \item The factors of variability that the error bars are capturing should be clearly stated (for example, train/test split, initialization, random drawing of some parameter, or overall run with given experimental conditions).
        \item The method for calculating the error bars should be explained (closed form formula, call to a library function, bootstrap, etc.)
        \item The assumptions made should be given (e.g., Normally distributed errors).
        \item It should be clear whether the error bar is the standard deviation or the standard error of the mean.
        \item It is OK to report 1-sigma error bars, but one should state it. The authors should preferably report a 2-sigma error bar than state that they have a 96\% CI, if the hypothesis of Normality of errors is not verified.
        \item For asymmetric distributions, the authors should be careful not to show in tables or figures symmetric error bars that would yield results that are out of range (e.g. negative error rates).
        \item If error bars are reported in tables or plots, The authors should explain in the text how they were calculated and reference the corresponding figures or tables in the text.
    \end{itemize}

\item {\bf Experiments compute resources}
    \item[] Question: For each experiment, does the paper provide sufficient information on the computer resources (type of compute workers, memory, time of execution) needed to reproduce the experiments?
    \item[] Answer: \answerYes{} 
    \item[] Justification: We describe it in Appendix~\ref{sec:hyperparameter_searches}.
    \item[] Guidelines:
    \begin{itemize}
        \item The answer NA means that the paper does not include experiments.
        \item The paper should indicate the type of compute workers CPU or GPU, internal cluster, or cloud provider, including relevant memory and storage.
        \item The paper should provide the amount of compute required for each of the individual experimental runs as well as estimate the total compute. 
        \item The paper should disclose whether the full research project required more compute than the experiments reported in the paper (e.g., preliminary or failed experiments that didn't make it into the paper). 
    \end{itemize}
    
\item {\bf Code of ethics}
    \item[] Question: Does the research conducted in the paper conform, in every respect, with the NeurIPS Code of Ethics \url{https://neurips.cc/public/EthicsGuidelines}?
    \item[] Answer: \answerYes{} 
    \item[] Justification: We reviewed the NeurIPS Code of Ethics and found no potential harms of our work.
    \item[] Guidelines:
    \begin{itemize}
        \item The answer NA means that the authors have not reviewed the NeurIPS Code of Ethics.
        \item If the authors answer No, they should explain the special circumstances that require a deviation from the Code of Ethics.
        \item The authors should make sure to preserve anonymity (e.g., if there is a special consideration due to laws or regulations in their jurisdiction).
    \end{itemize}

\item {\bf Broader impacts}
    \item[] Question: Does the paper discuss both potential positive societal impacts and negative societal impacts of the work performed?
    \item[] Answer: \answerYes{} 
    \item[] Justification: We discuss it in the problem statement in the Introduction and in the Conclusion.
    \item[] Guidelines:
    \begin{itemize}
        \item The answer NA means that there is no societal impact of the work performed.
        \item If the authors answer NA or No, they should explain why their work has no societal impact or why the paper does not address societal impact.
        \item Examples of negative societal impacts include potential malicious or unintended uses (e.g., disinformation, generating fake profiles, surveillance), fairness considerations (e.g., deployment of technologies that could make decisions that unfairly impact specific groups), privacy considerations, and security considerations.
        \item The conference expects that many papers will be foundational research and not tied to particular applications, let alone deployments. However, if there is a direct path to any negative applications, the authors should point it out. For example, it is legitimate to point out that an improvement in the quality of generative models could be used to generate deepfakes for disinformation. On the other hand, it is not needed to point out that a generic algorithm for optimizing neural networks could enable people to train models that generate Deepfakes faster.
        \item The authors should consider possible harms that could arise when the technology is being used as intended and functioning correctly, harms that could arise when the technology is being used as intended but gives incorrect results, and harms following from (intentional or unintentional) misuse of the technology.
        \item If there are negative societal impacts, the authors could also discuss possible mitigation strategies (e.g., gated release of models, providing defenses in addition to attacks, mechanisms for monitoring misuse, mechanisms to monitor how a system learns from feedback over time, improving the efficiency and accessibility of ML).
    \end{itemize}
    
\item {\bf Safeguards}
    \item[] Question: Does the paper describe safeguards that have been put in place for responsible release of data or models that have a high risk for misuse (e.g., pretrained language models, image generators, or scraped datasets)?
    \item[] Answer: \answerNA{} 
    \item[] Justification: We produce no harmful artifacts.
    \item[] Guidelines:
    \begin{itemize}
        \item The answer NA means that the paper poses no such risks.
        \item Released models that have a high risk for misuse or dual-use should be released with necessary safeguards to allow for controlled use of the model, for example by requiring that users adhere to usage guidelines or restrictions to access the model or implementing safety filters. 
        \item Datasets that have been scraped from the Internet could pose safety risks. The authors should describe how they avoided releasing unsafe images.
        \item We recognize that providing effective safeguards is challenging, and many papers do not require this, but we encourage authors to take this into account and make a best faith effort.
    \end{itemize}

\item {\bf Licenses for existing assets}
    \item[] Question: Are the creators or original owners of assets (e.g., code, data, models), used in the paper, properly credited and are the license and terms of use explicitly mentioned and properly respected?
    \item[] Answer: \answerYes{} 
    \item[] Justification: Yes, we credit all the datasets, models and methods used, and we are in compliance with their licenses.
    \item[] Guidelines:
    \begin{itemize}
        \item The answer NA means that the paper does not use existing assets.
        \item The authors should cite the original paper that produced the code package or dataset.
        \item The authors should state which version of the asset is used and, if possible, include a URL.
        \item The name of the license (e.g., CC-BY 4.0) should be included for each asset.
        \item For scraped data from a particular source (e.g., website), the copyright and terms of service of that source should be provided.
        \item If assets are released, the license, copyright information, and terms of use in the package should be provided. For popular datasets, \url{paperswithcode.com/datasets} has curated licenses for some datasets. Their licensing guide can help determine the license of a dataset.
        \item For existing datasets that are re-packaged, both the original license and the license of the derived asset (if it has changed) should be provided.
        \item If this information is not available online, the authors are encouraged to reach out to the asset's creators.
    \end{itemize}

\item {\bf New assets}
    \item[] Question: Are new assets introduced in the paper well documented and is the documentation provided alongside the assets?
    \item[] Answer: \answerYes{} 
    \item[] Justification: We provide the MIT license for our code. There are no other created assets.
    \item[] Guidelines:
    \begin{itemize}
        \item The answer NA means that the paper does not release new assets.
        \item Researchers should communicate the details of the dataset/code/model as part of their submissions via structured templates. This includes details about training, license, limitations, etc. 
        \item The paper should discuss whether and how consent was obtained from people whose asset is used.
        \item At submission time, remember to anonymize your assets (if applicable). You can either create an anonymized URL or include an anonymized zip file.
    \end{itemize}

\item {\bf Crowdsourcing and research with human subjects}
    \item[] Question: For crowdsourcing experiments and research with human subjects, does the paper include the full text of instructions given to participants and screenshots, if applicable, as well as details about compensation (if any)? 
    \item[] Answer: \answerNA{} 
    \item[] Justification: There was no crowdsourcing.
    \item[] Guidelines:
    \begin{itemize}
        \item The answer NA means that the paper does not involve crowdsourcing nor research with human subjects.
        \item Including this information in the supplemental material is fine, but if the main contribution of the paper involves human subjects, then as much detail as possible should be included in the main paper. 
        \item According to the NeurIPS Code of Ethics, workers involved in data collection, curation, or other labor should be paid at least the minimum wage in the country of the data collector. 
    \end{itemize}

\item {\bf Institutional review board (IRB) approvals or equivalent for research with human subjects}
    \item[] Question: Does the paper describe potential risks incurred by study participants, whether such risks were disclosed to the subjects, and whether Institutional Review Board (IRB) approvals (or an equivalent approval/review based on the requirements of your country or institution) were obtained?
    \item[] Answer: \answerNA{} 
    \item[] Justification: We did not use human subjects.
    \item[] Guidelines:
    \begin{itemize}
        \item The answer NA means that the paper does not involve crowdsourcing nor research with human subjects.
        \item Depending on the country in which research is conducted, IRB approval (or equivalent) may be required for any human subjects research. If you obtained IRB approval, you should clearly state this in the paper. 
        \item We recognize that the procedures for this may vary significantly between institutions and locations, and we expect authors to adhere to the NeurIPS Code of Ethics and the guidelines for their institution. 
        \item For initial submissions, do not include any information that would break anonymity (if applicable), such as the institution conducting the review.
    \end{itemize}

\item {\bf Declaration of LLM usage}
    \item[] Question: Does the paper describe the usage of LLMs if it is an important, original, or non-standard component of the core methods in this research? Note that if the LLM is used only for writing, editing, or formatting purposes and does not impact the core methodology, scientific rigorousness, or originality of the research, declaration is not required.
    \item[] Answer: \answerNA{} 
    \item[] Justification: We did not use LLMs for the core work, only for editing and code completion.
    \item[] Guidelines:
    \begin{itemize}
        \item The answer NA means that the core method development in this research does not involve LLMs as any important, original, or non-standard components.
        \item Please refer to our LLM policy (\url{https://neurips.cc/Conferences/2025/LLM}) for what should or should not be described.
    \end{itemize}

\end{enumerate}

\end{document}